\documentclass[authoryear,preprint,12pt,times]{elsarticle}
\usepackage[T1]{fontenc} 
\usepackage{amssymb}
\usepackage{amsmath}
\usepackage{hyperref}
\usepackage{xcolor}
\usepackage{subcaption}
\usepackage{caption}
\usepackage{graphicx}
\usepackage{multicol}
\usepackage{multirow}
\usepackage{pdflscape}
\usepackage{makecell}
\usepackage{booktabs}
\usepackage{fontawesome5}
\usepackage{pifont}
\usepackage{float}
\usepackage[absolute,overlay]{textpos}
\usepackage{boxedminipage}

\journal{}

\begin{document}

\begin{frontmatter}

\title{It is not always greener on the other side: Greenery perception across demographics and personalities in multiple cities}

\author[sec]{Matias Quintana}
\author[chk]{Fangqi Liu}
\author[fin]{Jussi Torkko}
\author[sec,doa]{Youlong Gu}
\author[doa]{Xiucheng Liang}
\author[doa]{Yujun Hou}
\author[doa]{Koichi Ito}
\author[doa]{Yihan Zhu}
\author[doa]{Mahmoud Abdelrahman}
\author[fin]{Tuuli Toivonen}
\author[chk]{Yi Lu}
\author[doa,dre]{Filip Biljecki\corref{cor}}

\affiliation[sec]{organization={Future Cities Lab Global, Singapore-ETH Centre},
            addressline={CREATE campus, \#06-01 CREATE Tower}, 
            city={Singapore},
            postcode={138602}, 
            country={Singapore}}
\affiliation[chk]{organization={Department of Architecture and Civil Engineering, City University of Hong Kong},
            city={Hong Kong},
            postcode={},
            country={Hong Kong SAR}}
\affiliation[fin]{organization={Digital Geography Lab, Department of Geosciences and Geography, University of Helsinki},
            city={Helsinki},
            postcode={},
            country={Finland}}
\affiliation[doa]{organization={Department of Architecture, National University of Singapore}, 
            addressline={4 Architecture Drive}, 
            city={Singapore},
            postcode={117566}, 
            country={Singapore}}      
\affiliation[dre]{organization={Department of Real Estate, National University of Singapore}, 
            addressline={15 Kent Ridge Dr}, 
            city={Singapore},
            postcode={119245}, 
            country={Singapore}}
\cortext[cor]{Corresponding author}

\begin{abstract} 
\begin{textblock*}{\textwidth}(3.8cm,-0.1cm) 
\begin{center}
\begin{footnotesize}
\begin{boxedminipage}{1\textwidth}
This is the Accepted Manuscript version of an article published by Elsevier in the journal \emph{Landscape and Urban Planning} in 2026, which is available at:\\ \url{https://doi.org/10.1016/j.landurbplan.2026.105618}\\ Cite as:
Quintana M, Liu F, Torkko J, Gu Y, Liang X, Hou Y, Ito K, Zhu Y, Abdelrahman M, Toivonen T, Lu Y, Biljecki F (2026): It is not always greener on the other side: Greenery perception across demographics and personalities in multiple cities. \textit{Landscape and Urban Planning} 271: 105618.
\end{boxedminipage}
\end{footnotesize}
\end{center}
\end{textblock*}

\begin{textblock*}{1.5\textwidth}(2.2cm,26cm)
{\tiny{\copyright{ }2026, Elsevier. Licensed under the Creative Commons Attribution-NonCommercial-NoDerivatives 4.0 International (\url{http://creativecommons.org/licenses/by-nc-nd/4.0/})}}
\end{textblock*}

Quantifying and assessing urban greenery is consequential for planning and development, reflecting the everlasting importance of green spaces for multiple climate and well-being dimensions of cities.
Evaluation can be broadly grouped into objective (e.g.,\ measuring the amount of greenery) and subjective (e.g.,\ polling the perception of people) approaches, which may differ -- what people see and feel about how green a place is might not match the measurements of the actual amount of vegetation.
In this work, we advance the state of the art by measuring such differences and explaining them through human, geographic, and spatial dimensions.
The experiments rely on contextual information extracted from street view imagery and a comprehensive urban visual perception survey collected from 1,000 people across five countries with their extensive demographic and personality information. 
We analyze the discrepancies between objective measures (e.g.,\ Green View Index (GVI)) and subjective scores (e.g.,\ pairwise ratings), examining whether they can be explained by a variety of human and visual factors such as age group and spatial variation of greenery in the scene.
The findings reveal that such discrepancies are comparable around the world and that demographics and personality do not play a significant role in perception.
Further, while perceived and measured greenery correlate consistently across geographies (both where people and where imagery are from), where people live plays a significant role in explaining perceptual differences, with these two, as the top among seven, features that influences perceived greenery the most.
This location influence suggests that cultural, environmental, and experiential factors substantially shape how individuals observe greenery in cities.
We also found that the spatial arrangement of greenery in the sight, rather than its proximity to the person, influences perception.
Our study provides a new understanding of the deep relationships between objective and subjective street-level greenery assessments, contributing to a more human-centric design of green urban environments.

\end{abstract}




\end{frontmatter}
\section{Introduction}\label{sec:intro}
Quantifying and assessing greenery is crucial for planning and development, and has been the subject of research for decades.
Particularly, green spaces are favored due to their benefits for climate mitigation~\citep{Fujiwara.20243ao, Fujiwara.2024}, physical~\citep{Mao.2025, gao2024a, lu2019c, lu2018a}, and mental~\citep{Jiang.2014xp, Wang.2025hf} well-being, and economic development~\citep{Wu.2022mb}.
Traditionally, researchers have assessed greenery with data collected from aerial or satellite platforms~\citep{Li.20152ird}, with Street View Imagery (SVI) becoming an increasingly popular urban data source for similar purposes~\citep{Kang.2020, Biljecki.2021}.
These images have enabled researchers and practitioners to gather objective~\citep{Ye.201911, He.2025di, Li.20152ird, ki2021a, Wu.2022mb, Li.2024g84, zhang2024f, Mao.2025, merdymshaeva2025, zhao2025}, and subjective~\citep{Yang.2009, Quintana.2025, Lefosse.2025} greenery measurements from the human-centric perspective.

One widely used objective measurement to assess eye level greenery is the Green View Index (GVI)~\citep{Yang.2009, Ye.201911, He.2025di, Li.20152ird, ki2021a, Wu.2022mb, Li.2024g84, zhang2024f, Mao.2025, merdymshaeva2025, zhao2025}, which complements remote sensing, specifically orthophotography-based metrics such as the Normalized Difference Vegetation Index (NDVI)~\citep{Helbich.2021, Huang.2025, He.2025di}.
When computed from SVI, GVI extends traditional greenery variables such as the number of street trees and parks areas as it can capture broader greenery, including also other aspects such as lawns and green walls~\citep{Ki.2021}.
Besides objective measurements such as GVI, SVI has been handy to measure subjective dimensions (e.g.,\ through surveys where people rate their perception of greenery).
Evaluating both of these has been important because people's impressions of how green a place feels can differ significantly from technical measurements such as vegetation coverage~\citep{Torkko.2023}.
This discord means that a place could be measured to be highly green but still feel less green to people, or vice versa.
These differences stem from the measurement mechanism; biological vision focusing in spatial forms, elements, and colours, compared to ratio of vegetation pixels~\cite{Cao.2025}; and are also driven by many factors that are inherent to human perception, such as personal preference, cultural background, and the visibility, density, and type of greenery~\citep{LIU2024128335}.

Research comparing subjective and objective greenery has been conducted using both in-situ and image-based approaches, though their findings are not always directly comparable. 
In in-situ work, SVI-derived GVI has been shown to underestimate participants' field-based impressions of greenery~\citep{Torkko.2023}, potentially reflecting the broader sensory and contextual cues present during on-site experience.
Image-based studies have typically relied on perceptual ratings or greenery manual selection directly from SVI.
For example,~\citet{Ye.201911} asked respondents to judge greenery levels and \citet{long2017, Suppakittpaisarn.2022} quantified perceived greenery through respondents' greenery manual segmentation. 
Both reported positive associations with objective metrics, but their analyses were limited to single-city or single-country contexts and did not examine the subjective–objective relationship further.
Beyond greenery,~\citet{Lefosse.2025} analyzed the subjective perception of biophilia on imagery across biomes and demographics, providing global insights on the benefits of eye contact with nature-based elements.
These location and demographic differences in subjective measurements are crucial for providing targeted context-aware urban solutions~\citep{Qi.2022}.

However, efforts to explore the relationships between objective and subjective metrics across locations and demographics, and to comprehensively explore other image-derived features, remain scarce.
The few that combined GVI with greenery perception do so at a single city, limiting understanding of whether their relationship differ across geographies and populations~\citep{Torkko.2023, He.2025di}.
The rare multi-city efforts across demographics focus on broader topics such as biophilia~\citep{Lefosse.2025} or focus solely on NDVI and GVI with no subjective measurements~\citep{Huang.2025}.
Our work addresses these gaps by exploring the relationship between objective and subjective metrics across multiple locations and diverse participants who have volunteered to divulge their thorough personal information for this research.

\begin{figure}[tbp]
    \centering
    \includegraphics[clip, trim= 0cm 7cm 16.5cm 0cm, width=1\linewidth]{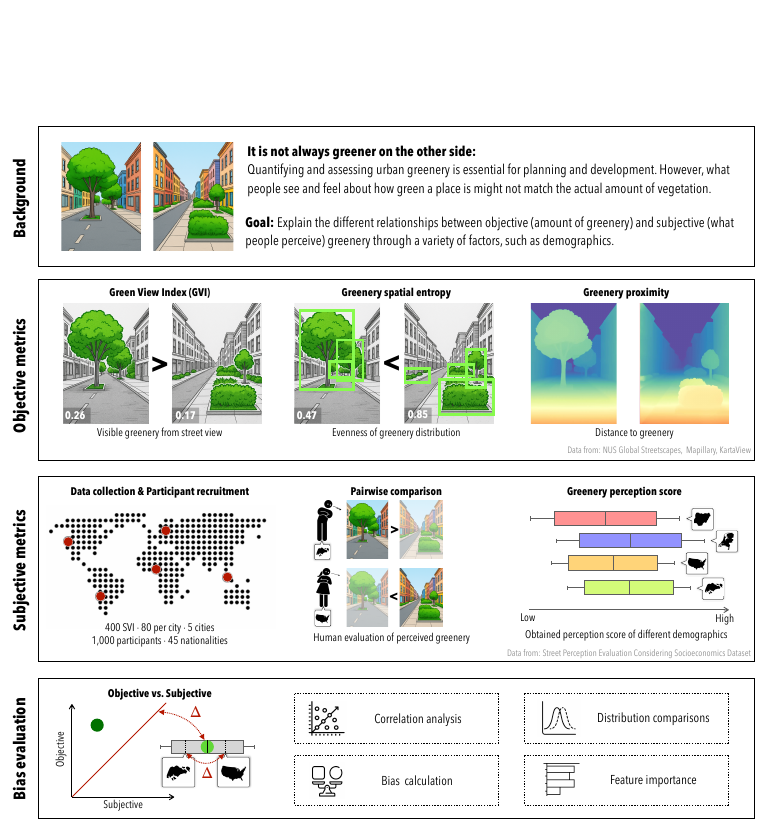}
    \caption{
    Workflow and methodology of this study: dataset identification, data preprocessing, and scores comparisons.
    We used the Street Perception Evaluation Considering Socioeconomic (SPECS) dataset due to its diverse samples and inclusion of a \textit{greenery} perceptual indicator~\citep{Quintana.2025}.
    We calculated green perception scores based on ratings for all 400 images, obtained the Green View Index (GVI) using the panoptic semantic segmentation models recommended in~\citep{Hou.2024}, and computed greenery spatial arrangement metrics based on the segmented greenery.
    We then compared these subjective, i.e., greenery perception, and objective, i.e., GVI, measurements in terms of their correlation and bias.
    Finally, as a last comparison, we looked into the vegetation spatial arrangement within images among the least and most green-rated images.
    }
    \label{fig:methodology}
\end{figure}

In this work, we utilize the comprehensive Street Perception Evaluation Considering Socioeconomics (SPECS) dataset, which is a large-scale global survey we designed and collected to support multiple lines of urban perception research~\citep{Quintana.2025}, to analyze and understand the relationship between commonly used objective and subjective methods used to evaluate greenery.
Among other information, the effort collected perception ratings on diverse but controlled imagery (i.e.,\ carefully and systematically curated and selected) from 1,000 people from around the world, together with a series of information about the participants and images.
First, we performed correlation and bias quantification analyses to thoroughly understand the relationship between objective and subjective greenery quantification at the street level.
Then, we looked deeper into the images by analyzing aspects such as the arrangement of greenery within the image and greenery proximity to the viewer.
Finally, we investigated which of these variables explains greenery perception the most by training a tree-based model using image-derived features and location information to predict greenery perception scores.
\autoref{fig:methodology} shows an overview of this work.
Through this comprehensive set of analyses of a multi-city dataset, we address three key questions that have not been investigated before at this scale and depth: a) How do subjective greenery evaluations relate to objective (GVI, vegetation arrangement, and proximity to the viewer) measurements? b) Which demographic factor most strongly influences greenery perception? c) How does the spatial arrangement of greenery affect subjective ratings?

\section{Literature review}\label{sec:lit-review}
\subsection{Subjective human perception of greenery}
Perceived greenery is shaped by both individual characteristics and broader geographic contexts. 
Studies have shown that demographic groups can experience and evaluate nature-based elements differently. 
For instance, \citet{Lefosse.2025} found that women preferred more nature-based elements than men, with additional variation attributable to city of residence.
Beyond stated preferences, other work has examined physiological and psychological responses to greenery.
\citet{Jiang.2014xp} demonstrated that tree-cover density yields diminishing stress-reduction benefits for men compared to women, while \citet{lottrup2013} reported similar gender-differentiated relationships between workplace greenery and stress. 
Age-related differences have also been observed, with green space immediately outside the home providing stronger developmental benefits for young girls than for boys~\citep{taylor2002}. 
Together, these findings highlight that perceptions of and responses to greenery are not uniform, but embedded within demographic, cultural, and place-based contexts.

Studies has employed a range of approaches to measure these perceptions.
Most studies evaluate greenery qualitative on-site~\citep{falfan2018, leslie2010, Sugiyamae9} or under controlled laboratory conditions~\citep{Aoki01121991}, often using street-level images or SVI~\citep{Yang.2009, Quintana.2025, Lefosse.2025, ogawa_evaluating_2024}.
In-situ assessments range from providing ratings from 0-100\%~\citep{falfan2018} to a Likert scale of different perceived green elements~\citep{leslie2010, Sugiyamae9}.
Image-based evaluations range from manual green segmentation~\citep{Yang.2009, long2017} to tracking physiological reactions to 3D videos~\citep{Jiang.2014xp}, to pair-wise comparison tasks~\citep{Quintana.2025, Lefosse.2025, ogawa_evaluating_2024}.
Compared with field-based assessments, image-based methods and online surveys are more time- and cost-effective~\citep{Kang.2020} and allow large-scale collection of perceptual data~\citep{Gu.2025, Salminen.2025}.

Although subjective evaluations are susceptible to human-induced biases~\citep{gupta2012}, they capture qualitative dimensions of greenery that quantitative approaches may overlook~\citep{leslie2010}.
Importantly, they help reveal demographic- and place-specific preferences, providing valuable insights for inclusive, context-aware urban design practices and strategies~\citep{Qi.2022}.

\subsection{Objective machine-derived perception of greenery}
Numerous indices and methods have been developed for computing the amount of greenery, such as remote sensing-based NDVI~\citep{He.2025di, Helbich.2021, Huang.2025} and point-cloud based methods~\citep{macfaden.2012, Yu.2016kla, labib2021, zieba-kulawik2021}.
However, methods relying on SVI, such as GVI~\citep{Ye.201911, He.2025di, Li.20152ird, ki2021a, Wu.2022mb, Li.2024g84, zhang2024f, Mao.2025, merdymshaeva2025, zhao2025, lu2019c, lu2018a} are predominantly used as an objective or computational approach to assess street and eye level greenery.
This method consists of calculating the visibility of greenery from a specific position~\citep{Yang.2009} by computationally extracting the ratio of vegetation elements' pixels over the entire image via color thresholding~\citep{Li.201569s, Li.20152ird, lu2019c, lu2018a} or semantic segmentation~\citep{Aikoh.2023, Hou.2024}.

Unlike remote sensing imagery, street-level imagery better captures what people experience~\citep{Ye.201911}, allowing GVI and other objective measurements to capture information that otherwise cannot be gathered from top-down perspectives~\citep{gaw2022, Li.2024g84, Helbich.2021}.
Additionally, it is straightforward to acquire and not very sensitive to quality issues, as even single (non-panoramic) imagery is sufficient~\citep{Biljecki.2023uda}.
This data also offers a wider global coverage and availability from proprietary (e.g., Google Street View or Baidu) and crowdsourced services (e.g., Mapillary or KartaView)~\citep{Biljecki.2021, Hou.2024}.

GVI has been found to be a relevant measure and predictor in the built environment --- it was used to examine the relationship between street greenery and socioeconomic profiles~\citep{Li.2024g84, Li.20152ird}, examine its relationship with overall physical activity~\citep{Mao.2025}, measure greenery cooling effect for heat stress~\citep{merdymshaeva2025}, and used as labeled data for GVI prediction models based on multi-source remote-sensing data~\citep{Ma.2025jf}.
Despite this adoption and ease of use, GVI faces challenges.

In relation to NDVI, studies have reported systematic differences when using GVI, although the direction and magnitude of these differences are context dependent. 
While \citet{Huang.2025} observed that GVI tended to yield lower greenery estimates than NDVI across their sampled cities,~\citet{Ye.201911} demonstrated that discrepancies between the two metrics are not consistently unidirectional but instead exhibit regional patterns, with GVI occasionally exceeding NDVI in certain urban contexts.
These variations are partly attributable to methodological differences between street-level and satellite-based imagery, including seasonal effects~\citep{zhao2025}, spatial bias~\citep{zhang2024f}, and sampling strategies~\citep{Huang.2025}.
In an attempt to overcome its shortcomings, researchers have developed more specialized versions of GVI to capture the full 360$^\circ$ view~\citep{Li.201569s} or the view from multi-story buildings using floor-specific adjustment~\citep{Yu.2016kla}.

\subsection{Objective meets subjective}
Although each type of measurement has its benefits and both aim to assess greenery quantitatively, fundamentally, they measure different things: objective measurements quantify the visible greenery numerically and consistently, while subjective measurements assess greenery based on people's perceptions, encapsulating both perceived quantity and quality.
Thus, recent attempts to understand the relationship between objective and subjective measurements have resulted in approaches combining both types of measurements to obtain a complete picture of urban greenery.
\citet{He.2025di} combined satellite imagery and SVI-based perceived safety to identify the relationship between greenery, through NDVI, and safety perception scores.
Using SVI, approaches range from biological-inspired measurements (i.e., landscape imagery, spatial forms, landscape elements, and color features) for subjective perception scores prediction~\citep{Cao.2025}, color-emotion analysis for measuring the positive impact of blue and green spaces~\citep{Chen.2025}, to quantifying biophilia based on image-based elements (objective measures) and pair-wise comparisons (subjective measures)~\citep{Lefosse.2025}.
Outside of greenery, urban design subjective preferences, and street physical features, both obtained from SVI, complement each other when used to explain housing prices~\citep{Qiu.2022}.
These approaches highlight opportunities where both types of measurements complement each other or serve to understand the other.
Nevertheless, these comparisons and hybrid methods, despite being done at a granular level within a city, have been constrained to single-city studies in places like Houston (USA)~\citep{He.2025di}, Tianjin (China)~\citep{Chen.2025}, Helsinki (Finland)~\citep{Torkko.2023}, and Stockholm (Sweden)~\citep{merdymshaeva2025}, overlooking one of the main limitations of subjective-based measurements: its location- and people-specific nature.
Current multi-city approaches either look solely at NDVI and GVI with no subjective measurements~\citep{Huang.2025} or focus on broader topics such as biophilia~\citep{Lefosse.2025}.

In this paper, we address these gaps by conducting a multi-city, multi-population experiment to evaluate the relationships and potential systematic differences between objective greenery measures (GVI, vegetation arrangement, and proximity to the viewer) and subjective greenery perceptions. 
We further examine how demographic and personality characteristics -- derived from our large global survey -- help explain these patterns.

\section{Methods}\label{sec:methods}

\subsection{Dataset}
We performed a survey that comprises pairwise comparison ratings of street view imagery from five different cities (80 images per city, a total of 400) contributed by 1,000 participants from 5 countries (200 participants from each country), for which we collected detailed demographic and personality information.
The cities where the images come from are balanced worldwide: Santiago (Chile), Amsterdam (Netherlands), Abuja (Nigeria), Singapore (Singapore), and San Francisco (USA), and the imagery was initially obtained from the NUS Global Streetscapes dataset~\citep{Hou.2024}.
The survey participants came from the same five countries.
As we mentioned in~\citep{Quintana.2025}, where we introduced this dataset (called SPECS -- Street Perception Evaluation Considering Socioeconomics), we chose these cities to following a three-tier filtering process: (1) continental representation to ensure global coverage, (2) availability in established datasets (MIT PP2~\cite{Dubey.2016} and NUS Global Streetscapes~\citep{Hou.2024}) to enable comparative analyses, and (3) consideration of cultural, urban form, and socioeconomic diversity within these constraints.

These criteria were essential for evaluating our core hypothesis that urban perception --- across many perceptual indicators including \textit{green} --- varies across demographic groups and across cities with different spatial and cultural characteristics.
The five cities therefore intentionally span diverse built forms, socioeconomic profiles, and cultural settings (Figure 1; Supplementary Information in~\cite{Quintana.2025}). 
This diversity is a necessary condition for testing perceptual robustness across heterogeneous environments.

From these cities, we sample images following the image download process from~\cite{Hou.2024} and download all available street view images within a 2.4 km$^2$ square in the city center (with the exception of Abuja, for which we manually selected a similarly sized but more developed area).
We then filtered images based on available metadata (e.g.,\ clear weather and image taken during the day); visual metrics such as visual complexity; and image composition ratios of roads, vegetation, cars, buildings, and sky)~\cite{Quintana.2025}.
None of the final 400 selected images showed snow or ground cover attributable to weather conditions.
By keeping the quality and properties of imagery consistent, we minimise the potential bias of various aspects of imagery on perception.

The information we collected from 1,000 participants from the five countries include gender, age, nationality, country and city of residence, length of stay in the said city, annual income level, number of household members, race and ethnicity, and Big Five Inventory personalities (\textit{extraversion}, \textit{agreeableness}, \textit{conscientiousness}, \textit{neuroticism}, and \textit{openness}).
The dataset contains the participants' pairwise comparison ratings for the commonly used six indicators (\textit{safe}, \textit{lively}, \textit{wealthy}, \textit{beautiful}, \textit{boring}, and \textit{depressing}) proposed in the MIT Place Pulse 2.0 dataset (PP2)~\citep{Dubey.2016}, which have been widely and consistently adopted in the literature~\cite{Cui.2023c2p, Kang.2023, meir_understanding_2020, hidayati_how_2020, ogawa_evaluating_2024, liang_evaluating_2024}.
Using them also facilitates comparative analysis with previous and future studies.
In addition, we also included four new indicators of \textit{live nearby}, \textit{walk}, \textit{cycle}, and \textit{green} which are better aligned with livability, active mobility, and sustainability study goals.
To support this research, we introduced this last mentioned indicator and was prompted to users in the form of ``Which place looks \textbf{greener}?''.
A snippet of the dataset is shown in \autoref{tab:snippet-specs} and the final dataset contained 50,000 pairwise comparisons ratings across all participants and perceptual indicators (i.e.,\ 1,000 participants $\times$ 50 pairwise comparisons per survey on which participants rated imagery from all five cities).

We calculated machine-derived metrics from the images, such as the amount of greenery, greenery arrangement, and greenery proximity, and computed the \textit{greenery} perception scores from the pairwise comparison ratings based on the participants' location.

\begin{landscape}
\begin{table}[]
    \centering
    \begin{tabular}{ccccc|ccc|cccc}
    \hline
    \multicolumn{5}{c}{\makecell{Demographic data\\(multiple choice)}} & 
    \multicolumn{3}{c}{\makecell{Big Five personality score\\(Bounded in [1, 7])}} & 
    \multicolumn{4}{c}{\makecell{Pairwise comparisons\\(multiple choice: left, right, equal)}}\\
    \makecell{Participant\\ID} & Gender & Age group & \makecell{Country of\\residence} & \dots & Extraversion & \dots & Openness & Indicator & \makecell{Left\\image ID} & \makecell{Right\\image ID} & Choice\\
    \hline
    146 & Female & 40-49 & Nigeria & \dots & 3 & \dots & 2.5 & safe & 307 & 99 & right \\ 
    146 & Female & 40-49 & Nigeria & \dots & 3 & \dots & 2.5 & green & 88 & 284 & equal \\ 
    \multicolumn{5}{c}{\vdots} & \multicolumn{3}{c}{\vdots} & \multicolumn{4}{c}{\vdots}\\
    31 & Male & 21-29 & Singapore & \dots & 2.5 & \dots & 2.5 & safe & 397 & 148 & left \\
    31 & Male & 21-29 & Singapore & \dots & 2.5 & \dots & 2.5 & green & 20 & 340 & right \\
    \multicolumn{5}{c}{\vdots} & \multicolumn{3}{c}{\vdots} & \multicolumn{4}{c}{\vdots}\\
    \end{tabular}
    \caption{
    Snippet of the SPECS dataset containing its three main sections: participants' demographic data, Big Five personality scores, and pairwise comparisons ratings.
    Each participant completed an onboarding survey that included the demographic and personality questions (10 questions each) and 50 pairwise comparisons ratings (5 questions per indicator) of the 400 randomized images.
    The complete list of questions and dataset details are available in~\cite{Quintana.2025}.
    }
    \label{tab:snippet-specs}
\end{table}
\end{landscape}

\subsection{Objective measurements}
\subsubsection{Amount of greenery: Green View Index}
GVI is the visibility of greenery from a specific position~\citep{Yang.2009} and can be calculated as shown in the equation below~\citep{Li.20152ird}.
\begin{equation}\label{eq:gvi}
G V I=\frac{\text { Number of Green pixels }}{\text { Number of Total pixels }}
\end{equation}
With the index ranging from 0 to 1, when a lot of greenery is visible from a specific position, the index value is high.
To compute GVI, instance-based image analysis is first used to segment green vegetation.
This segmentation is commonly done with spectral information, i.e., using the different pixel values from the RGB bands~\citep{Li.201569s, Li.20152ird}, or with semantic segmentation machine learning models~\citep{Aikoh.2023, Hou.2024}.
We compared both approaches and, similar to~\citep{Torkko.2023}, chose the semantic segmentation approach as it is less sensitive to small variations in pixel values, such as lighting conditions or dimly/brightly lit greenery~\citep{pietikainen1996, batlle2000}.
\autoref{app:fig:spectral-vs-semantic} shows an example of a pair of images from the SPECS dataset with their respective GVI based on spectral and semantic segmentation (middle and right column, respectively).
We ran the panoptic semantic segmentation, i.e., instance and semantic segmentation, inferences using the ZenSVI Python package~\citep{Ito.2025}, which uses a model trained on the Mapillary Vistas version 1.2 dataset~\citep{Neuhold.201716f} as the backend.
We computed the GVI by including areas segmented as both `vegetation' and `terrain' categories. 
Following \citep{torkko2025} and their GVI-based methodological recommendations, we retained `terrain' segments to explicitly capture natural elements within the GVI calculation. 
To ensure consistency, we manually validated `terrain' segments by examining the segmented street view imagery and including only those areas that represented grass surfaces or vegetated patches.

\subsubsection{Greenery distribution: Spatial entropy}
We quantified the distribution of vegetation within the image.
While existing work in urban studies has used visual complexity as a measure of entropy, e.g.,\ whether the image is composed predominantly of one element or multiple ones~\citep{Mayer.20239w9, Yap.2023dysn}, this computation does not take into account the spread of the elements.
Calculating the entire image entropy would provide an overall vegetation proportion without accounting for spatial arrangements.
For example, an urban scene with vegetation evenly distributed across the image and an urban scene with vegetation concentrated on one side would have the same entropy values.
Thus, as we are particularly interested in how the identified vegetation (e.g.,\ trees, bushes, grass patches) is spatially distributed in an image, we opted for computing the spatial entropy as a measure of the spatial heterogeneity, or randomness, of vegetation within an image.

For a given sliding window $W$ of size $s \times s$, the spatial entropy is defined using the Shannon entropy formula (\autoref{eq:shannon-entropy}) where $p$ is the proportion of vegetation pixels in the window $W$ (\autoref{eq:proportion}) and $1 - p$ represents the proportion of non-vegetation pixels. 
To obtain the overall spatial entropy for the image, we first generate a vegetation mask and then compute the mean entropy across all valid windows (\autoref{eq:mean_entropy}), where $N$ is the total number of sliding windows across the image and $H(W_i)$ is the entropy computed for the $i$-th window.
Spatial entropy is bounded in $[0, 1]$, and a high value indicates a more mixed and uncertain distribution, while lower entropy suggests more uniformity within the image.

\begin{equation}\label{eq:shannon-entropy}
H(W) = - p \log_2 p - (1 - p) \log_2 (1 - p)
\end{equation}

\begin{equation}\label{eq:proportion}
p = \frac{\sum W}{s^2}    
\end{equation}

\begin{equation}\label{eq:mean_entropy}
H_{\text{mean}} = \frac{1}{N} \sum_{i=1}^{N} H(W_i)    
\end{equation}

\begin{figure}[tbp]
\centering
    \begin{subfigure}{\linewidth}
        \centering
        \includegraphics[clip, trim=0cm 2cm 12cm 2cm, width=0.7\linewidth]{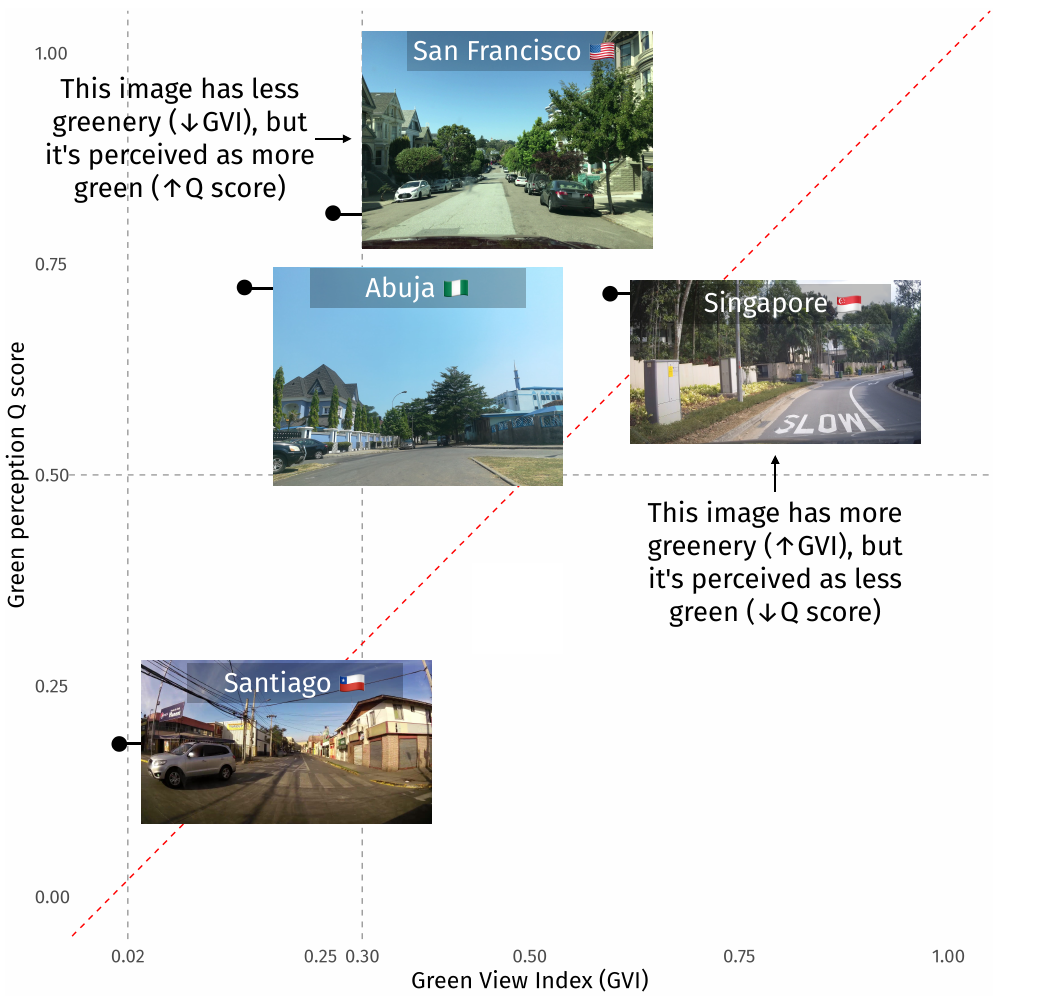}
        \caption{
        Discrepancy between subjective and objective notions of urban greenery.
        A red dashed line shows the identity function and vertical lines are shown as thresholds for imagery with some vegetation (GVI $= 0.3$), very little vegetation (GVI $ = 0.02$), and perception Q score of 0.5.
        }
    \label{fig:metrics-example-gvi-qscore}
    \end{subfigure}
    \hfill
    \begin{subfigure}{\linewidth}
        \centering
        \includegraphics[clip, trim=0cm 5cm 5.5cm 6.5cm, width=0.9\linewidth]{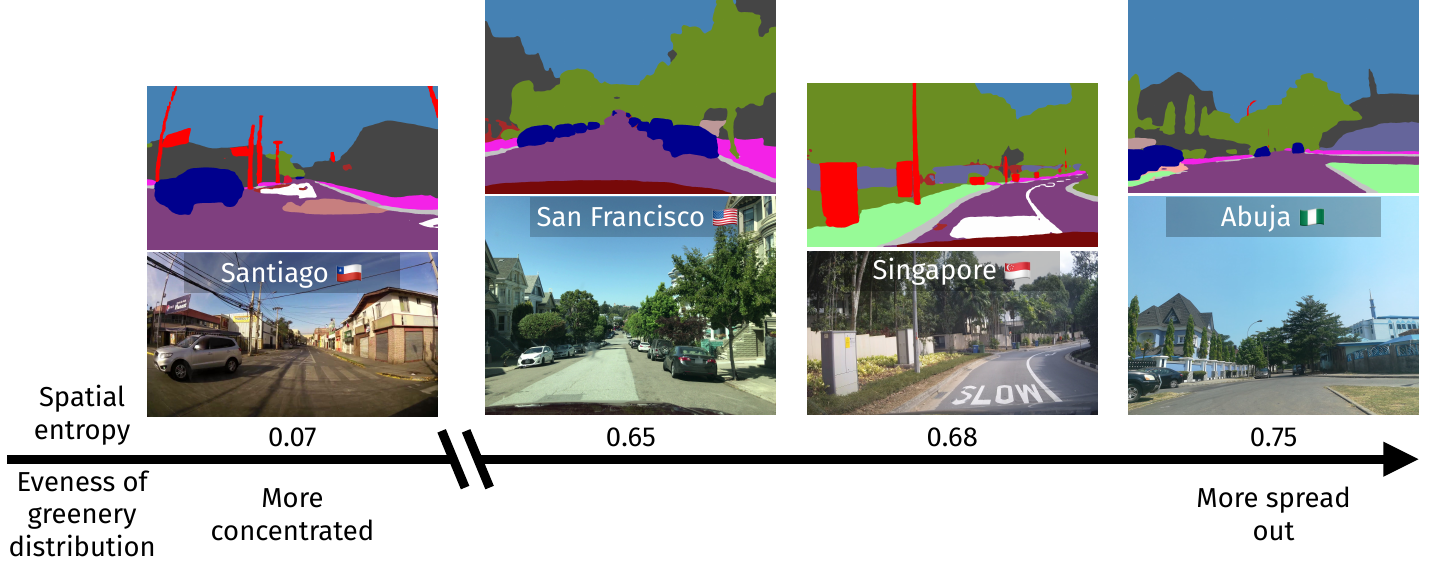}
        \caption{
        Higher values of spatial entropy indicate vegetation distributed throughout the urban scene, and lower values indicate vegetation concentrated in specific locations.
        At comparable spatial entropy values (between 0.68 and 0.65), the SVI from Singapore has a higher GVI than the one from San Francisco (0.60 and 0.27, respectively).
        }
        \label{fig:metrics-example-spatial}
    \end{subfigure}
    \caption{
    Four images from the SPECS dataset~\citep{Quintana.2025} arranged based on their (a) \textit{green} perception Q scores (y-axis) and their Green View Index (GVI) values (x-axis), and their (b) spatial entropy. 
    }
    \label{fig:metrics-example}
\end{figure}

Due to the different image sizes, a characteristic of crowdsourced SVI owing to heterogeneous equipment used~\citep{Hou.2024}, we opted for a window size based on the images' dimensions, i.e.,\ relative window size, instead of a fixed dimension.
We performed a sensitivity analysis (\autoref{app:fig:entropy-sensitivity}) on window size fractions ranging from 0.10 to 1.00 with 0.05 increments and found the maximum average value at a window size fraction of 0.45.
This peak entropy window size represents the scale at which the image exhibits the most spatial variation between vegetation and non-vegetation.
For the dataset, this selected window size fraction resulted in squared sliding windows from $480\times480$ to $921\times921$ pixels. 
\autoref{fig:metrics-example} shows an example of five images from the dataset arranged based on their respective aggregated \textit{green} perception Q scores (y-axis) and GVI values (x-axis) (\autoref{fig:metrics-example-gvi-qscore}, and spatial entropy values (\autoref{fig:metrics-example-spatial}).

\subsubsection{Greenery proximity: Distance to the viewer}
In addition to quantifying the spatial entropy of vegetation within the image, we also measured its proximity to the viewer. This metric is crucial because identical GVI values can arise from vastly different spatial configurations, potentially leading to distinct subjective perceptions. As noted in classic Gestalt psychology, the principle of proximity is fundamental to visual perceptual grouping, determining how discrete elements are organized into a coherent whole~\citep{Wagemans2012}. For instance, a scene dominated by large canopy structures in the distance may yield the same GVI as one with smaller vegetation in the immediate foreground. However, purely pixel-based metrics fail to distinguish between these scenarios. By incorporating the distance variable, we aim to disentangle these spatial attributes, allowing us to examine whether human perception is driven more by the immediate visual impact of nearby vegetation or the extensive coverage of distant landscapes.

To estimate the distance of vegetation, we utilized the Depth Anything V2 model~\citep{yang2024d}.
This advanced monocular depth estimation algorithm has already demonstrated strong performance in urban domains such as street-view analysis and greenery perception~\citep{ito2025, ma2025}. 
Trained on a combination of high-quality synthetic images and over sixty million pseudo-labeled real images, the model delivers high accuracy and robust generalization, making it well-suited for analyzing street-view scenes without requiring ground-truth depth annotations.  

We calculated vegetation distances using the same masks employed for GVI calculation, i.e.,\ masks containing pixels classified as vegetation and green terrain from the panoptic semantic segmentation inference. 
These masks were overlaid on the depth maps to isolate vegetation pixels. 
For each image, we then derived two proximity indicators: \textit{absolute depth} (the mean distance of vegetation pixels to the observer) and \textit{relative depth} (the depth of vegetation compared to other scene elements). 
Together, these measures capture whether greenery is predominantly near or far from the viewer and extend the GVI by incorporating a proximity dimension into the assessment.

\subsection{Subjective measurements: Perception ratings}
The numerical perception scores were computed using the Strength of Schedule (SOS) method and are referred to as Q scores, ranging from $[0, 10]$ and reflecting the magnitude of the perception.
The dataset also provides the Q scores based on demographics, i.e., Q scores calculated using ratings from only a specific demographic group and based on the image and participants' locations.
Another widely used numerical scoring method for pairwise comparisons in urban visual perception studies is the Trueskill ranking system~\citep{Kang.2023, Qiu.2025}.
Both scores, Q scores and Trueskill scores~\citep{herbrich_trueskill_2006}, inherently depend on the ratings considered for their calculations.
Thus, rather than being an absolute measure of an image's perceptual dimensions, these scores are only comparative; they are only valid within the rating pool in which they were calculated.

We compared the relative scores for the \textit{green} perceptual indicator and found mostly similar distributions.
\autoref{app:fig:qscore-trueskill-qq} shows a Quantile-Quantile plot of Trueskill scores and Q scores based on location-pair (i.e.,\ country-city pair) groups based on the images' and participants' country and city, respectively.
While there are some discrepancies around lower scores (e.g.,\ values above the diagonal line in \autoref{app:fig:qscore-trueskill-qq}, Chile-Abuja location pair), we opted for Q scores as we used them in the original SPECS dataset~\citep{Quintana.2025} and recommended in recent relevant work~\citep{Gu.2025}.
We also followed the SPECS dataset preprocessing criteria of only retaining images with at least four pairwise comparisons in each grouping context, as this was the reported average comparison in PP2~\citep{Dubey.2016}.
Finally, we normalized the computed Q scores into the $[0, 1]$ range to match the range of GVI.

\subsection{Analyzing the measurements relationships}
To systematically evaluate the agreement between subjective \textit{green} perception Q scores and objective GVI values, we employed a multi-step statistical approach that accounts for the different measurement scales and distributional properties of these metrics.
First, we performed a Pearson's correlation analysis, following previous objective-subjective greenery studies~\citep{Suppakittpaisarn.2022, Torkko.2023}, to quantify the linear relationship between perception and measurement. 
However, correlation alone does not reveal systematic biases or the magnitude of disagreement between metrics.
Therefore, following the analysis in~\citep{Suppakittpaisarn.2022}, we conducted a Bland-Altman analysis, which visualizes the arithmetic difference, i.e.,\ perception Q score - GVI, against the average of both metrics for each image.
This approach identifies whether disagreements are consistent across the measurement range or vary systematically with the magnitude of greenness.
Given that GVI represents an absolute metric while perception Q scores are relative measures, we normalized GVI values to ensure comparable scales. 
Then, to quantitatively confirm our visual assessments, we applied the non-parametric Wilcoxon signed-rank test. 
We chose this test because the perception Q scores exhibited non-normal distributions across different participants and image groups. 
The test compares the distribution of arithmetic differences, i.e.,\ perception Q score - normalized GVI, against zero, testing the null hypothesis $H_\theta:$ the median difference equals zero.
Rejection of this hypothesis indicates that one metric is consistently higher or lower than the other, revealing systematic measurement bias.

We also investigated whether the spatial characteristics of greenery within images influence green perception ratings. 
Specifically, we examined greenery distributions using spatial entropy measures and proximity analysis to test whether perceived greenness relates not only to the amount of greenery but also to its spatial arrangement. 
Images were divided into two groups based on their \textit{green} perception Q scores: those within the 25$^{th}$ ($\leq$Q1, perceived as less green) and those within the 75$^{th}$ ($\geq$Q3, perceived as more green) among their respective image-participant location pairs.
We applied the Mann-Whitney U test, a non-parametric alternative that does not assume normality, to compare spatial arrangements metrics between these groups. 
This analysis tests whether urban scenes perceived as highly green (\textit{green} perception Q score $\geq$Q3) exhibit systematically different greenery spatial patterns -- such as more dispersed vegetation (higher spatial entropy) or different peripheral distribution -- compared to scenes perceived as less green (\textit{green} Q score $\leq$Q1). 
Understanding these spatial-perceptual relationships could inform urban design strategies that optimize not just the quantity but also the arrangement of green elements to maximize positive greenery perception.

Finally, we trained a Random Forest model to predict \textit{green} perception Q scores using both participant characteristics, image-derived features, and perception Q scores for the remaining indicators.
The input features included participants' place of residence (whose pairwise comparisons generated the Q scores); along with image-based measures: Sky View Index, Green View Index, spatial entropy, and Shannon entropy; and the perception Q scores, calculated based on participants' place of residence, for the remaining nine indicators: \textit{safe}, \textit{lively}, \textit{wealthy}, \textit{beautiful}, \textit{boring}, \textit{depressing}, \textit{live nearby}, \textit{walk}, and \textit{cycle}.

Features such as greenery proximity and image location were excluded due to their limited relevance to the our modeling objectives. 
Greenery proximity distributions could not be meaningfully reduced to single scalar metrics without substantial information loss, unlike the spatial entropy measures, which effectively capture spatial patterns in single values.
We also excluded the image location (i.e.,\ the city where the SVI was captured) because it functions largely as a proxy identifier; including it would risk data leakage by enabling the model to exploit city-level differences rather than the street-level visual features that this analysis aims to isolate.

Our modeling goal is not to compare greenery levels across cities but to examine how different participant groups (by country) respond to the same set of urban scenes based on their visual characteristics. 
Random Forest was selected for its interpretability and suitability for tabular data.
To ensure robust and geographically balanced evaluation, we applied an 80-20 train-test split stratified by image-participant location combinations.

After computing the perception Q scores, based on the participants' locations and a combined location `All', for all 10 indicators on each image we retained scores with at least four pairwise comparisons, matching exiting thresholds~\citep{Quintana.2025} and other urban visual datasets~\citep{Dubey.2016}
The final dataset resulted in 1,903 images or around 79\% of the maximum 4,800 images (i.e.,\ 400 images $\times$ (5 cities $+$ 1 combined location)).
Of these, over 297 images were rated by participants from single locations (i.e.,\ cities), with 1,522 images used for training and 381 for testing. 
Model hyperparameters were optimized using 5-fold cross-validation on the training set. 
Detailed hyperparameter settings and training procedures are provided in~\autoref{app:tab:model-training}.

Although multicollinearity might be expected among image-derived features given their shared visual and spatial characteristics, its impact in Random Forest is minimised.
Because of its algorithm design, Random Forests are relatively robust to multicollinearity as each tree is train on random subset of features, reducing the chances of correlated features being used together.
Nevertheless, we further assess the model's performance with conditional permutation feature importance to account for potential multicollinearity.
Unlike standard permutation importance where a feature is dropped independently to all others, conditional permutation feature importance~\citep{strobl2008} permutes a feature while preserving its correlation structure with other features.
This permutation analysis prevents inflated or deflated importance metrics and produces unbiased importance estimates in a multicollinearity setting.

\section{Results}\label{sec:results}
The results show that perceived greenery and GVI are moderately related but not interchangeable: correlations are often medium to moderate across locations, yet perception scores tend to exceed GVI values.
Beyond quantity, the spatial arrangement and image-derived elements such as Sky View Index emerge as important drivers of perception, underscoring that how greenery is arranged within the streetscape matters alongside how much of it is present.

\subsection{Correlation analysis}
\begin{figure}[tbp]
\centering
    \begin{subfigure}{\linewidth}
        \centering
        \includegraphics[width=1\linewidth]{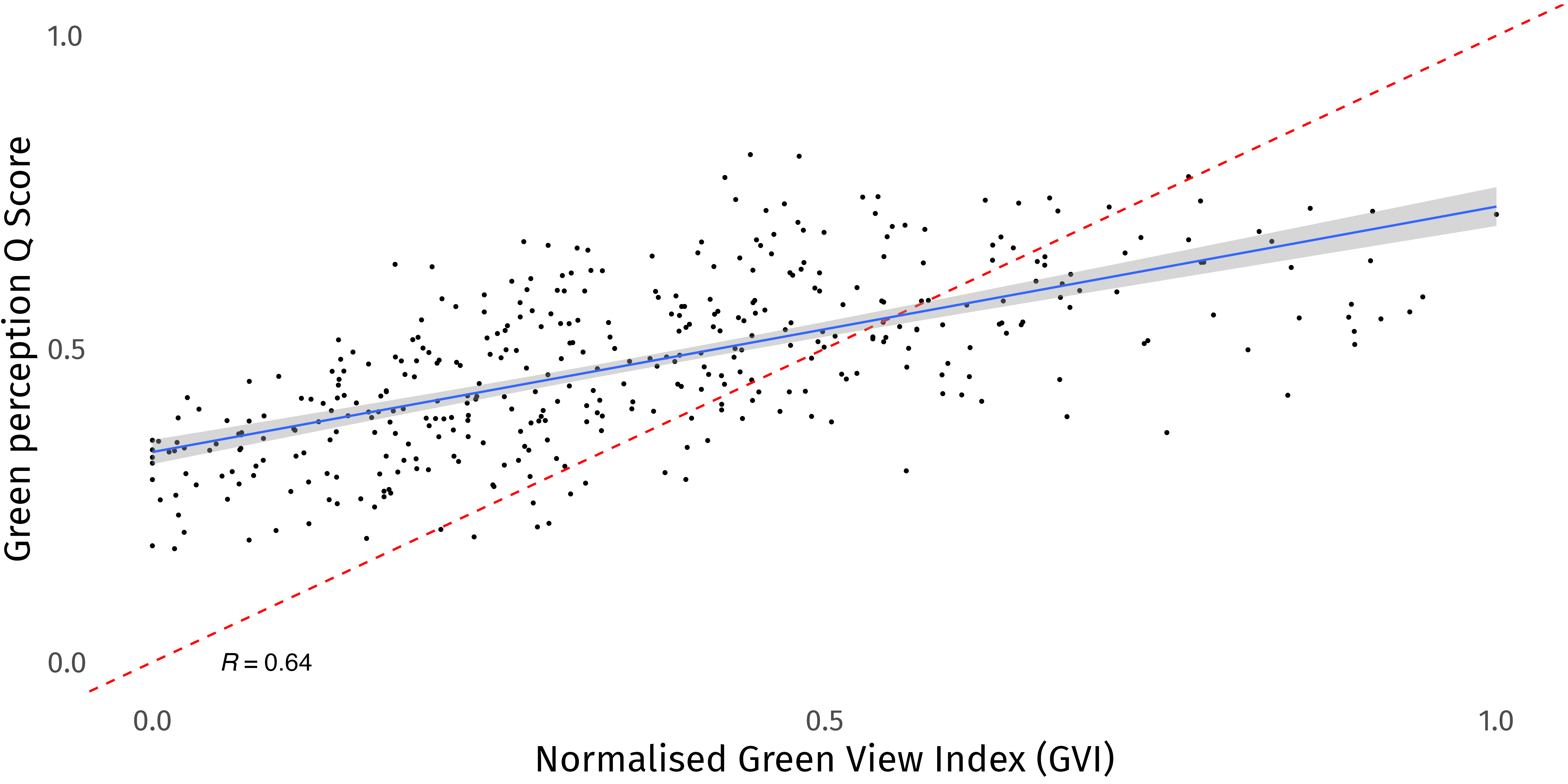}
        \caption{
        Pearson correlation scatter plot between the \textit{green} perception Q scores and normalised Green View Index (GVI) with the linear model in blue and an identity function as a red dashed line.
        The correlation is significant ($p<0.5$) and all 400 images ($n = 400$) are used.
        }
    \label{fig:green-gvi-overall-corr}
    \end{subfigure}
    \hfill
    \begin{subfigure}{\linewidth}
        \centering
        \includegraphics[width=1\linewidth]{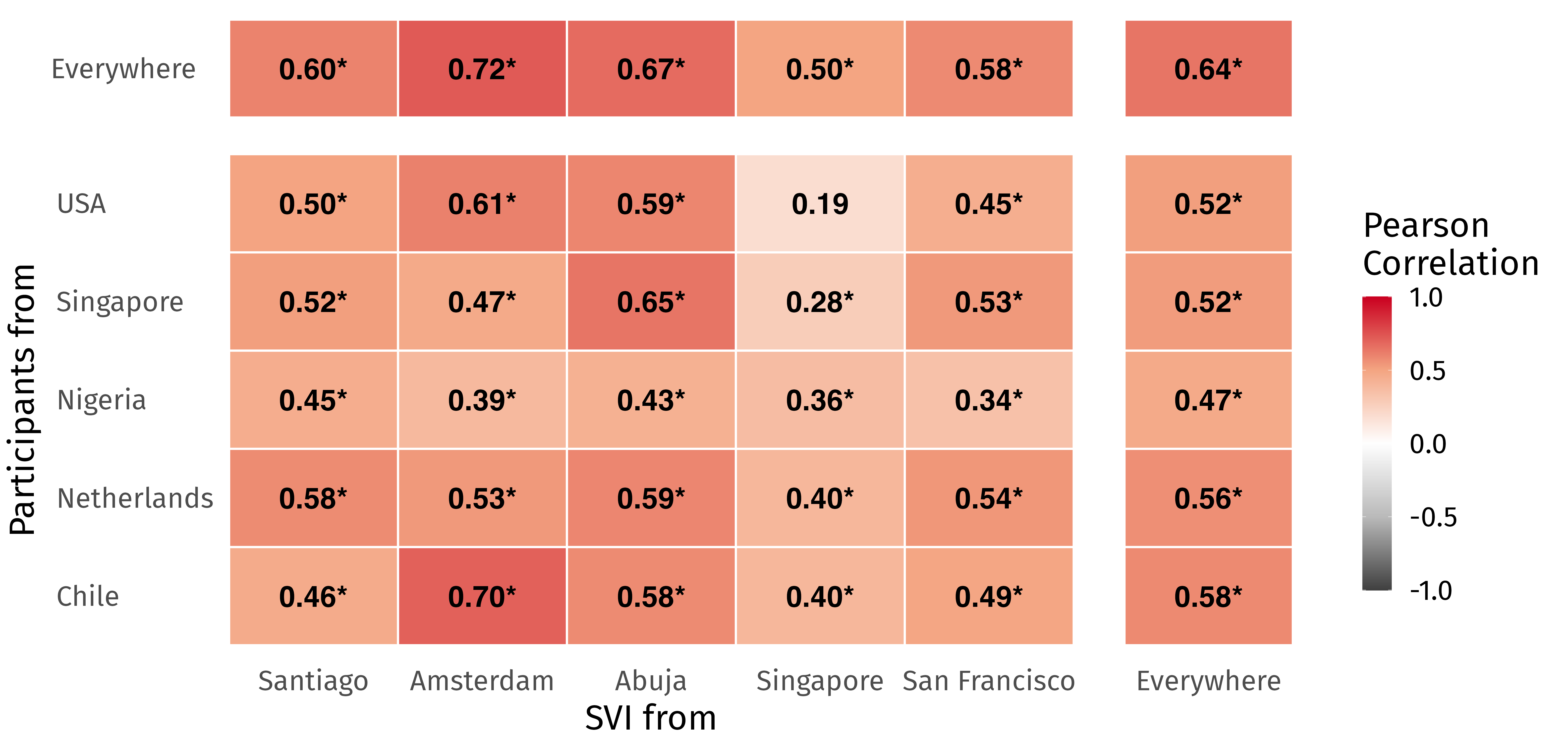}
        \caption{
        Pearson correlation heatmap between the \textit{green} perception Q scores and Green View Index (GVI) of images grouped by their location (x-axis) as well as all combined imagery (right-most column), rated by participants from different countries (y-axis).
        Most correlations are significant (bolded with a *$p<0.05$) and with an R-value $>0.50$.
        Each location pair (country-city pair) has at least 54 images with more than four pairwise comparisons ($n\geq54$).
        The top row presents a summarised overview of the correlation of all participants rating images from particular locations and the value in the top right represent the correlation for all combinations in the dataset.
        }
        \label{fig:green-gvi-corr-heatmap}
    \end{subfigure}
    \caption{
    Pearson correlation between the \textit{green} perception Q scores and Green View Index (GVI) using all images and responses (a) and grouped by the images' and participants location (b).
    }
    \label{fig:green-gvi-corr}
\end{figure}

\autoref{fig:green-gvi-corr} shows the overall correlations between \textit{green} perception Q scores and normalised GVI (\autoref{fig:green-gvi-overall-corr}) and based on the images' and participants' location (\autoref{fig:green-gvi-corr-heatmap}).
The analysis reveals moderate correlations overall and weak correlations within location pairs.
The overall correlation is significant with an R-value of $0.64$ and out of the 25 image-participant location pairs, all but one correlation is statistically significant, and in half of them (12), the R-value is $\ge 0.50$.
Most of the highest but still moderate correlations (R-value ranging from $0.43$ to $0.65$) are found on images from Abuja rated by participants across all countries, with the highest correlation (R-value of $0.70$) on images from Amsterdam rated by participants from Chile.
Weaker correlations (R value ranging from $0.19$ to $0.40$) are found on images from Singapore rated by participants across all countries.
More specifically, the two weakest combinations are: participants from USA rating greenery from imagery in Singapore (0.19) and participants from Singapore rating greenery in images from there (0.28), while the two strongest pairs are participants from Chile rating greenery from imagery in Amsterdam (0.70) and participants from Singapore rating greenery from imagery in Abuja (0.65).

A more detailed Pearson's correlation analysis is presented as scatter plots in \autoref{fig:app:green-gvi-corr-scatter_location} and \autoref{fig:app:green-gvi-corr-scatter}.
Unlike \autoref{fig:green-gvi-corr-heatmap}, in the latter figure, images are grouped solely based on either the images' location or the participants' location, not both.
The analysis reveals relatively strong correlations between \textit{green} perception Q scores and GVI.
In \autoref{fig:app:green-gvi-corr-scatter}, correlations average R $\approx 0.62$ for ratings given by all participants to imagery from each city (top row), and R $\approx 0.55$ for ratings given by participants from each country to all imagery (bottom row).

Following \citet{Jiang.2014xp}, who found an inverse relationship between tree cover density and stress reduction for men at around 30\% tree coverage, we examined regions with GVI $<0.3$. in detail. 
We identified two distinct regions of interest. 
The green region captures imagery with moderate vegetation (GVI $<0.3$) that is perceived as moderately green (\textit{green} perception Q score $> 0.5$).
The yellow region captures imagery with very little vegetation (GVI $<0.02$, more than ten times lower than the green region GVI threshold) that is perceived as not particularly green (\textit{green} perception Q score $< 0.5$).

For the green region, substantial variation exists across location pairs. 
In \autoref{fig:app:green-gvi-corr-scatter_location}, between 10\% and 50\% of images meet these criteria, with the lowest percentage for Santiago imagery rated by participants from Nigeria and the highest for Abuja imagery rated by participants from the Netherlands. 
When images are grouped by location (\autoref{fig:app:green-gvi-corr-scatter}, top row), the range is 9\% to 41\%, with Santiago showing the lowest and Abuja the highest percentages. 
However, when grouped by participants' location (\autoref{fig:app:green-gvi-corr-scatter}, bottom row), the distribution is more consistent, with approximately one-fourth (24\% to 27\%) of images meeting these criteria across all five countries.

The yellow region shows different patterns. 
No imagery from Abuja or Singapore, and almost none from San Francisco, falls within this region regardless of participant location (\autoref{fig:app:green-gvi-corr-scatter_location}). 
In contrast, 13\% to 17\% of Amsterdam imagery consistently meets these criteria. 
When examining all participants' ratings (\autoref{fig:app:green-gvi-corr-scatter}, top row), the percentage varies dramatically by city location (0\% in Singapore and Abuja, 19\% in Amsterdam), while ratings by participants from specific countries show consistently low percentages (4\% to 5\%) across all cities (\autoref{fig:app:green-gvi-corr-scatter}, bottom row).

\subsection{Bias evaluation}
\begin{figure}[tbp]
    \centering
    \includegraphics[width=0.97\linewidth]{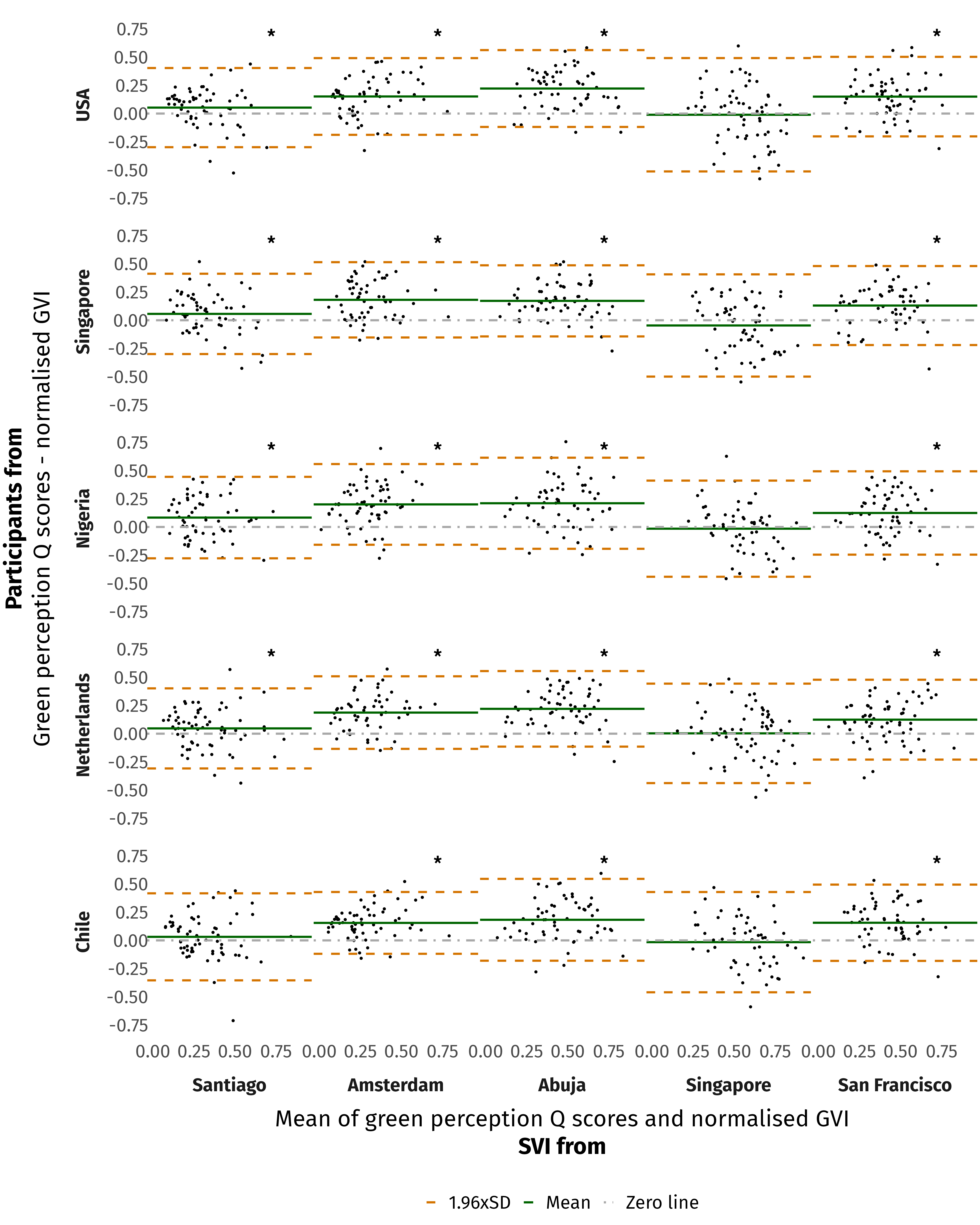}
    \caption{
    Bland-Altman plot comparing differences between \textit{green} perception Q Scores and normalized Green View Index (GVI) values across different location pairs (country-city pairs). 
    The plot shows the difference between measurements (y-axis) against their mean (x-axis) for all images and participants' location pairs, with green horizontal lines representing the mean difference, dashed orange lines indicating $\pm$1.96 standard deviations, and a dotted gray line at zero.
    Each location pair has more than 54 images with more than four pairwise comparisons ($n\geq54$).
    Location pairs with significant differences between \textit{green} perception Q score and normalized GVI are shown (*$p<0.05$).
    }
    \label{fig:bland-altman-location}
\end{figure}

Besides the correlation between \textit{green} perception Q scores and GVI values, and following the analysis based on images' and participants' locations, we developed a Bland-Altman plot or difference plot (\autoref{fig:bland-altman-location}) to confirm the agreement between the two measurements.
Using the Wilcoxon signed-rank test, we found the difference between \textit{green} perception Q scores and normalized GVI values is predominantly different from zero ($p<0.05$), suggesting the subjective and objective scores differ significantly across all location pairs, with a few noticeable exceptions.
There are no significant measurement differences in imagery from Chile rated by participants from Santiago, nor in all imagery from Singapore across all participants ratings. 
This suggests a better alignment between participants respective \textit{green} perception Q scores and these urban scenes' GVI.
For the remaining image-participant location pairs, the direction of this difference is visualized in~\autoref{fig:bland-altman-location} and results grouped instead by participants' and images' location separately in~\autoref{fig:app:bland-altman}.
These Bland-Altman plots show the vast majority of score differences, i.e.,\ \textit{greenery} perception Q score - normalized GVI, within two Standard Deviations (SD) of the mean (dashed orange line) and slightly more than have of them above the zero line (dotted gray line).
This result suggests that, in most cases, \textit{green} perception Q scores tends to be consistently higher than GVI values for the same urban scenes.

\subsection{Greenery distribution}

\begin{figure}
    \centering
    \begin{subfigure}{\linewidth}
        \centering
        \includegraphics[width=\linewidth]{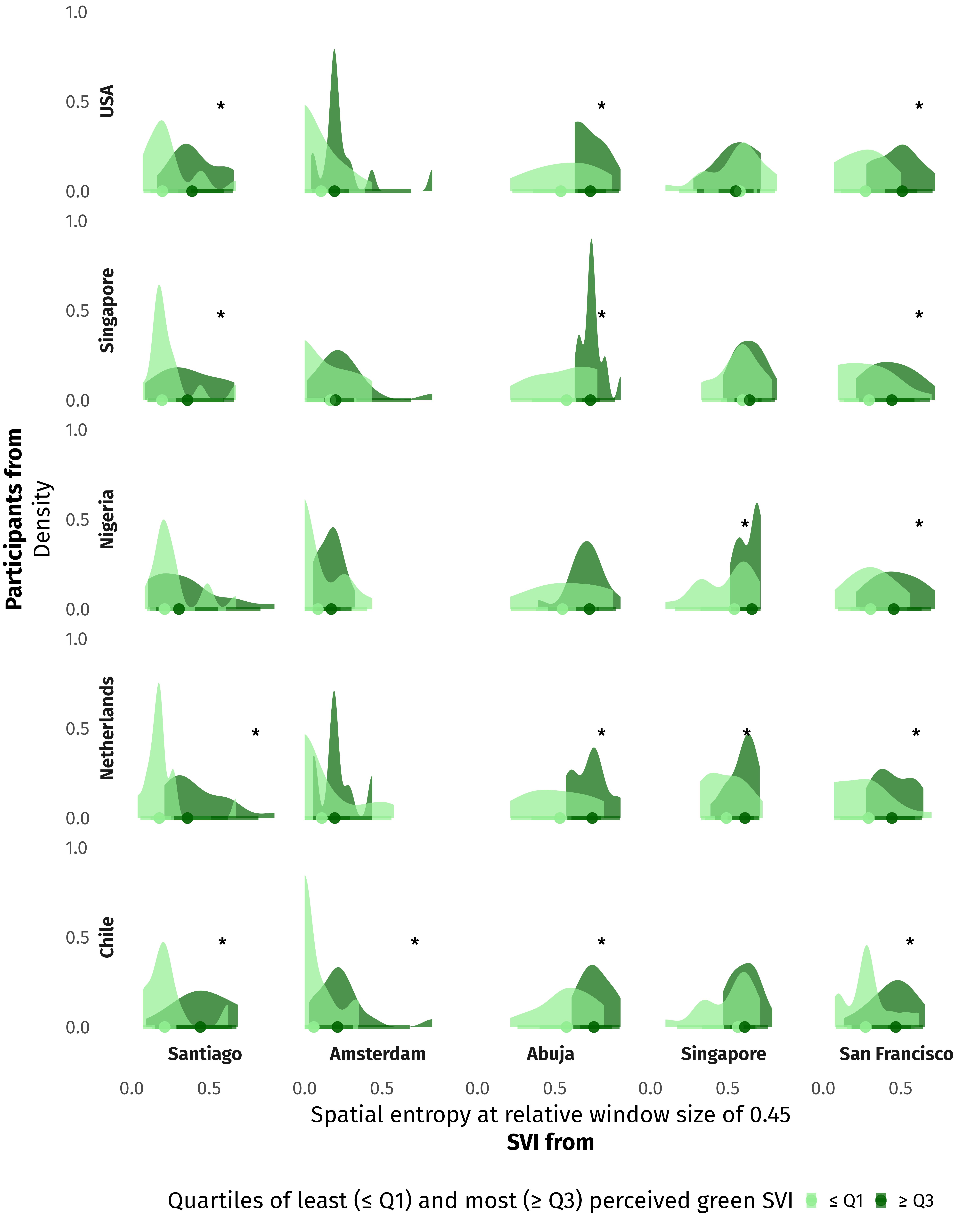}
        \caption{
            Greenery spatial entropy distributions.
        }
    \label{fig:spatial-entropy}
    \end{subfigure}
\end{figure}

\begin{figure}
    \ContinuedFloat
    \centering
    \begin{subfigure}{\linewidth}
    \centering
    \includegraphics[width=\linewidth]{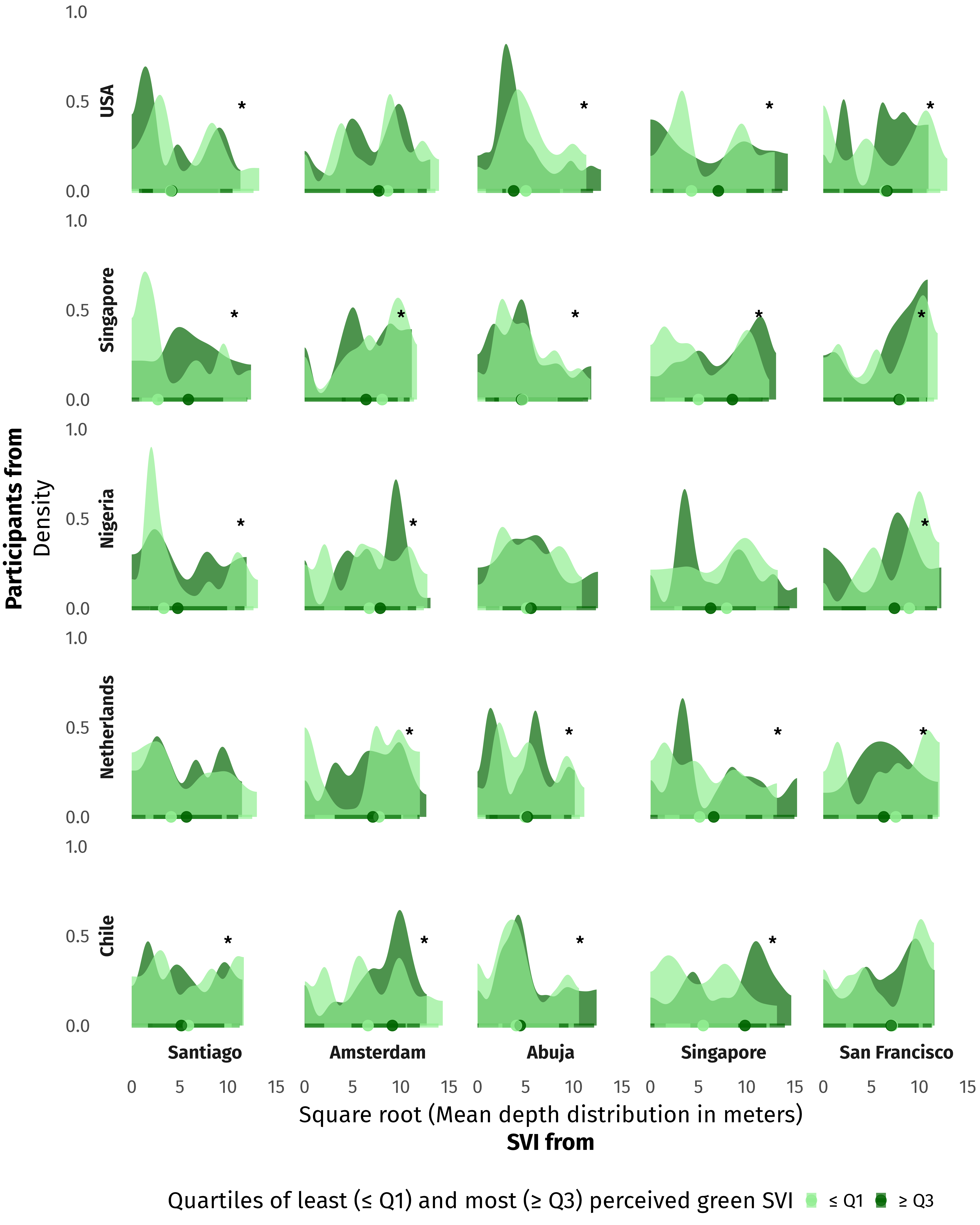}
    \caption{
        Greenery absolute depth distributions (in square root for visualization purposes).
    }
    \label{fig:depth-groupby-Qscores}
    \end{subfigure}
    \caption{
    Greenery distributions and statistical comparison of street view imagery (SVI) within the 25$^{th}$ ($\leq$Q1 in lighter green) or within the 75$^{th}$ ($\geq$Q3 in darker green) percentile of their \textit{green} perception Q scores.
    \textit{Green} perception Q scores are calculated based on the images' and participants' location pairs (country-city pairs),  with the median represented by a filled circle.
     Each location pair has more than 14 images with more than four pairwise comparisons ($n\geq14$).
     Significance thresholds *$p<0.05$.
    }
    \label{fig:vegetation-distributions}
\end{figure}

\autoref{fig:vegetation-distributions} shows the greenery spatial entropy (a) and greenery depth (b) distributions with significant differences in more than half (16 out of 25) and almost all (20 out of 25) location pairs, respectively.
On one hand, greenery spatial entropy values in images perceived as most green ($\geq$Q3) were significantly higher (\autoref{fig:spatial-entropy}) than in the images perceived as least green (darker green filled circle on the right side of the light green filled circle).
This suggests that images perceived as very green typically have greenery distributed throughout the urban scene rather than concentrated in specific locations.
On the other hand, greenery depth distributions showed wider spread and multiple peaks across all location pairs.
Among the 20 location pairs with statistically significant differences in mean depth distances, eight showed higher mean distances in the least perceived \textit{green} images ($\leq$Q1) while 12 showed higher mean distances in the most perceived \textit{green} images ($\geq$Q3) (\autoref{fig:depth-groupby-Qscores}).

\subsection{Greenery perception prediction}
\begin{figure}
    \centering
    \includegraphics[width=1\linewidth]{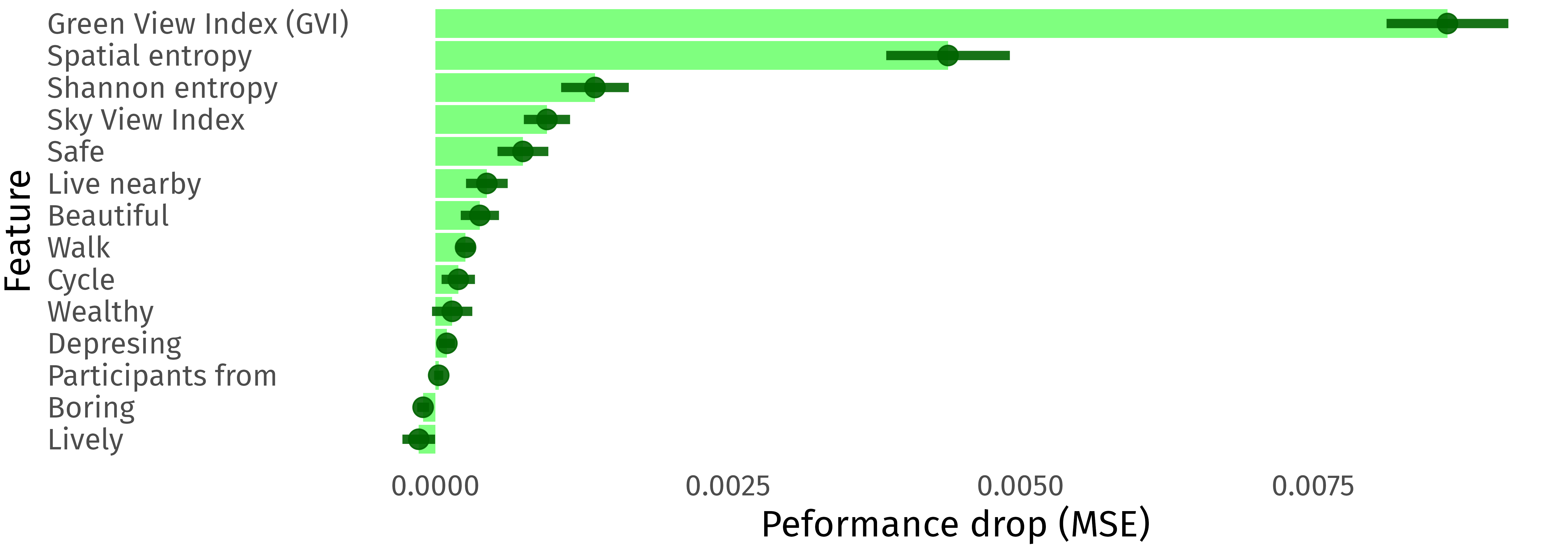}
    \caption{
    Conditional permutation feature importance analysis of the best performing Random Forest model.
    Mean and standard deviation of performance degradation (in Mean Squared Error (MSE), x-axis) when the values of a given feature (y-axis) are randomly shuffled while preserving its correlation structure with other features. 
    }
    \label{fig:rf-permutation}
\end{figure}

The best-performing Random Forest model was selected from 50 candidates with diverse hyperparameters (\autoref{app:tab:model-training}). 
The model achieved low Mean Squared Error (MSE) and moderate predictive performance (i.e.,\ 0.0191 MSE and 0.49 R$^2$, respectively) with comparable performance between 5-fold cross-validation and the test set, indicating good generalization without overfitting.
To identify which features most strongly explain greenery perception, we conducted a conditional permutation feature importance analysis. 
This method measures the performance drop (increase in MSE) when individual feature values are randomly shuffled while preserving its correlation structure with other features, thereby quantifying each feature's contribution to model predictions while accounting for multicollinearity. 
\autoref{fig:rf-permutation} presents the mean performance drop and standard deviation across multiple permutations for each feature.
The results confirm that GVI is the strongest predictor of \textit{green} perception Q scores. 
Additionally, image-derived features (spatial entropy, Shannon entropy, and Sky View Index) are among the most important features, followed by the perception Q scores for the \textit{safe} perceptual indicator.
Other positive perceptual indicators, such as \textit{live nearby}, \textit{beautiful}, \textit{walk}, \textit{cycle}, and \textit{wealthy}, provide a marginal contribution while the remaining perceptual indicators and the participants' location have a negligible contribution to the prediction model.
Complementing \autoref{fig:rf-permutation}, \autoref{tab:influence} shows an summary of the participants-based (demographics and personalities) and image-based (machine-derived metrics) features and their influence, or lack of it, in perceived greenery.

\begin{table}[htbp]
    \centering
    \begin{tabular}{p{4cm}lc}
        \toprule
        Group & Category / Variable & Influence \\ 
        \midrule
        People's profiles & \textbf{Demographics}\\
        & \quad Gender & \ding{55} \\
        & \quad Age group & \ding{55} \\
        & \quad Annual household income & \ding{55} \\
        & \quad Education level & \ding{55} \\
        & \quad Race and ethnicity & \ding{55} \\
        & \quad City of residence & \ding{52} \\
        & \textbf{Personalities}\\
        & \quad Extraversion & \ding{55} \\
        & \quad Agreeableness & \ding{52} \\
        & \quad Conscientiousness & \ding{55} \\
        & \quad Neuroticism & \ding{55} \\
        & \quad Openness & \ding{55} \\
        \midrule
        Image & \textbf{Machine-derived}\\
        & \quad Green View Index (GVI) & \ding{52} \\
        & \quad Spatial entropy & \ding{52} \\
        & \quad Proximity to viewer & \ding{55} \\
        & \quad Sky View Index & \ding{55} \\
    \bottomrule
    \end{tabular}
    \caption{
    Summary of demographics (gender, age group, annual household income, education level, race and ethnicity, and city of residence), personalities (extraversion, agreeableness, conscientiousness, neuroticism, and openness), image-derived features (Green View Index (GVI), spatial entropy, proximity to viewer, and Sky View Index), and perceptual indicators (\textit{safe}, \textit{lively}, \textit{wealthy}, \textit{beautiful}, \textit{boring}, \textit{depressing}, \textit{live nearby}, \textit{walk}, and \textit{cycle}) and their influence in perceived greenery evaluation (in Q scores) in street view imagery. 
    \ding{52} = influence, \ding{55} = no influence.
    }
    \label{tab:influence}
\end{table}

\section{Discussion}\label{sec:discussion}
\subsection{How little is too little? Perceived versus measured greenery}
As \autoref{fig:green-gvi-corr-heatmap} shows, perceived greenery and GVI values demonstrate strong agreement across nearly all images' and participants' location pairs.
This strong agreement is underscored by GVI being the most influential feature when predicting \textit{green} perception Q scores (\autoref{fig:rf-permutation}).
However, participants often overestimated greenery compared to objective measurements (\autoref{fig:bland-altman-location}), a finding that aligns with previous studies~\citep{Suppakittpaisarn.2022, Torkko.2023}.
\citet{Huang.2025} found an underestimation of GVI when comparing to NDVI, attributing the discrepancy to differences in data collection methods, such as sparse sampling points to match satellite-imagery with street view imagery, and analyzable image aspects. 
In contrast, our study uses the same data source, i.e.,\ SVI, for both subjective and objective measurements, suggesting that alternative sources of bias drive the observed differences.
One potential explanation lies in the fundamental nature of our measurements: GVI provides absolute greenery values, while perception Q scores are inherently relative measures. 
As noted in our previous work~\citep{Quintana.2025}, relative scoring methods like Q scores and TrueSkill scores depend entirely on the specific rating pool used for their calculation. 
Consequently, these scores are only valid within their original rating context, i.e., within the set of images used.

Another possible, and complementary, explanation relates to the inherently nuanced nature of subjective evaluation.
When assessing greenery, participants may have incorporated contextual cues, such as spatial arrangement or perceived distance, beyond the visible amount of vegetation. 
This contextualization suggests that subjective greenery perception, unlike objective measures such as GVI, reflects a more holistic assessment informed by multiple contextual factors.
Our correlation analysis aligns with this interpretation, showing systematic differences in how different populations perceive greenery.
Imagery from Singapore consistently yielded the lowest correlations with GVI across all participant groups (\autoref{fig:green-gvi-corr-heatmap}), while participants from the USA exhibited the lowest correlations when rating imagery from any city. 
For example, the differences between normalized GVI and \textit{green} perception Q scores for imagery from Singapore were not statistically significant, regardless of participants' location (\autoref{fig:bland-altman-location}) even when all participants' responses were combined (\autoref{fig:app:bland-altman}, top row).

This finding suggests that both the urban context being evaluated and the cultural background of the evaluator influence the relationship between objective and subjective greenness measures.
These patterns likely reflect location-specific perceptual baselines developed through extended urban exposure, as the location of the imagery is among the top three most influential features for predicting \textit{green} perception Q scores (\autoref{fig:rf-permutation}).
Given that the majority of participants have lived in their respective cities for more than five years~\citep{Quintana.2025}, their local urban characteristics may have become internalized reference points. 
For instance, participants from San Francisco and Santa Clara (USA) may not perceive dense urban scenes from Singapore as particularly green when compared to their urban greenery.

We noticed specific examples in three images -- one each from Abuja, Singapore, and San Francisco -- that were consistently rated as highly green (\textit{green} perception Q score $> 0.5$) by all participants regardless of their residence, despite having low greenery coverage (GVI $< 0.3$) (\autoref{fig:metrics-example-gvi-qscore}, middle left image). 
These findings demonstrate that greenery perception can transcend measurable vegetation, likely incorporating other natural elements, built environment features, or contextual cues that contribute to an overall sense of `greenery'.
These results also align with research on biophilia -- the positive emotional and sensory effects experienced when humans interact with nature. 
While \citet{Lefosse.2025} found that people perceive nature differently depending on their geographic context, they confirmed that greenery remains the dominant factor in positive environmental perception. 
Our findings extend this understanding by demonstrating that perceived greenery operates as a more complex construct than objective vegetation metrics suggest, incorporating cultural, contextual, and possibly aesthetic factors that merit further investigation in urban planning and design applications.

\subsection{Maybe it is not greener on the other side}
Looking into spatial greenery placement, our analysis of greenery location reveals distinct patterns in how spatial arrangements influences greenery perception.
Images perceived as most \textit{green} ($\geq$Q3) tend to feature greenery distributed throughout the scene rather than concentrated in specific areas (\autoref{fig:spatial-entropy}).
In contrast, greenery proximity to the viewer showed a context-dependent relationship with \textit{green} perception, demonstrating divergent patterns across location pairs (\autoref{fig:depth-groupby-Qscores}). For instance, observers (e.g., from the USA) viewing Abuja associated closer greenery with higher scores (aligning with the ``proximity effect'' described in Section 3.2.3), whereas perceptions of Singaporean imagery often favored greater depths, likely due to continuous green corridors that maintain visual impact even at a distance. To validate this finding, we conducted additional analysis comparing \textit{green} perception scores between images with close greenery (median depth $\leq$Q1) and distant greenery (median depth $\geq$Q3) (\autoref{fig:app:qscore_by_depth_location}). The perception rating distributions showed substantial overlap, indicating that greater distance does not inherently diminish perceived greenness. These findings nuance prior observations that distance has limited effect on perception~\cite{han2021}, suggesting instead that its role varies across different urban contexts.

Consistent with our \textit{green} perception Q scores predictive modeling, spatial entropy emerged as the most influential factor after GVI, with other image-derived features ranking just below (\autoref{fig:rf-permutation}).
Notably, greenery perception diverged from positive perceptual dimensions such as beauty or liveliness, which showed marginal and negligible predictive contribution, respectively.
This disconnection may reflect the framing of the task: participants were not explicitly asked to rate how beautiful or lively the greenery in the urban scene is, but to rate the scene as a whole.
We hypothesize that greenery judgments follow Gestalt principles, where vegetation is perceived holistically based on its overall spatial arrangements rather than its proximity to the viewer. 
This could explain why dispersed vegetation across the scene consistently received higher greenery ratings, even when located further from the camera.

\subsection{Demographic and personality differences}
We did not find significant differences in greenery perceptions across demographic groups (gender, age group, income, education level, and race and ethnicity)~\cite{Quintana.2025}.
This suggests that these demographic characteristics plays no role in perceiving greenery and that studies may be largely generalisable globally.
Personality also has no influence, except in a few specific combinations, e.g.,\ there were differences between people with high and low agreeableness personality trait scores in participants from Singapore~\citep{Quintana.2025}.
Moreover, despite our dataset being comprehensive (with respect to the state of the art), it still does not have enough responses per participant in nested demographics subgroups (e.g.,\ responses from men in the 21-29 age group) to perform exhaustive demographic interaction analyses that are statistically significant.
Our analysis revealed that imagery location and participants' geographic background (i.e.,\ city of residence) significantly influence greenery perception, as the relationship between this perceptual indicator and objective greenery measurements varies by location.
As such, ignoring residents' responses or using non-residents' responses for greenery subjective evaluation could lead to incorrect assessments.

\begin{figure}
    \centering
    \includegraphics[clip, trim=0cm 7.8cm 0cm 0cm, width=\linewidth]{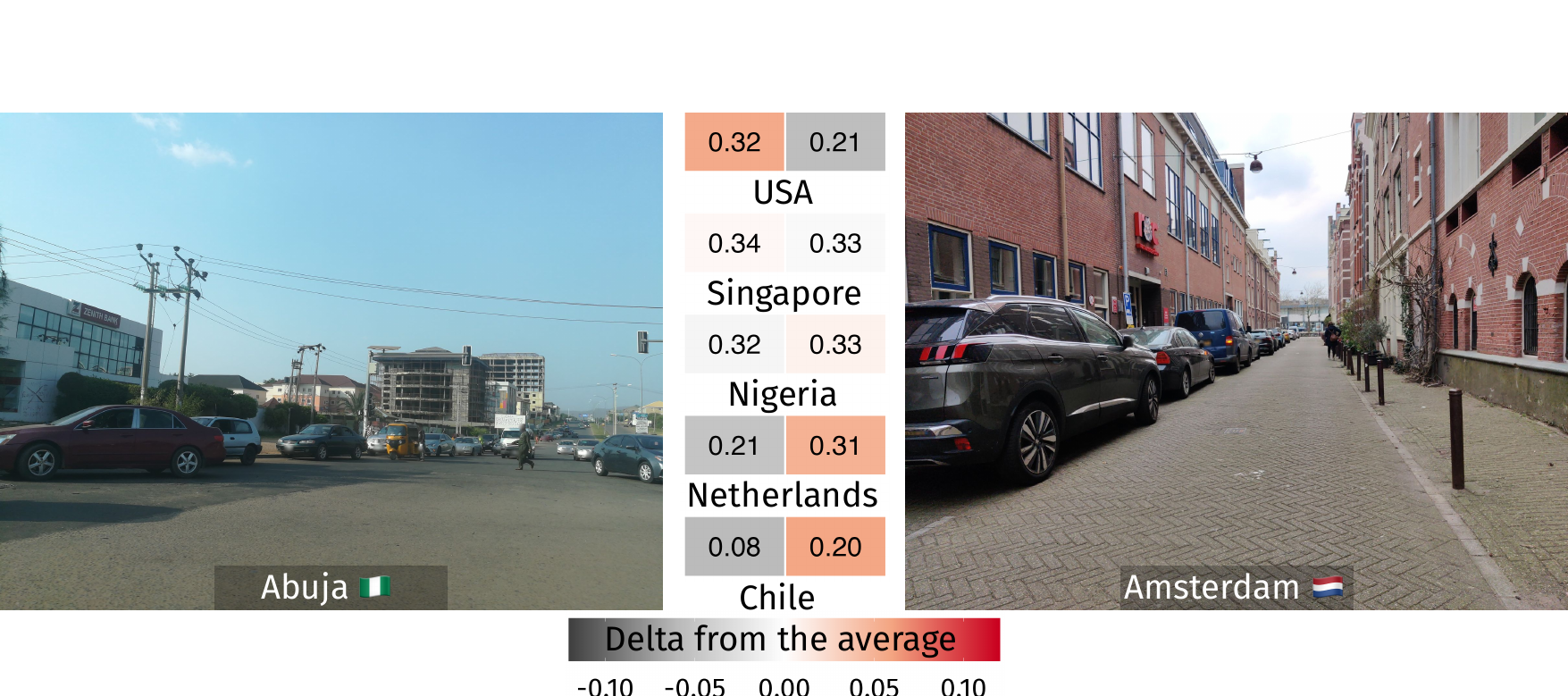}
    \caption{
    Example of different \textit{green} perception Q scores from two scenes from Abuja (left) and Amsterdam (right) between participants by their location.
    Both images have little greenery (GVI of $0.03$ and $0.04$, respectively) and are perceived as not too green (\textit{green} perception Q score $<$ 0.5).
    The individual street view image score, given by participants from a given country, is shown at the center of the figure next to the respective scene.
    The color gradient is calculated based on the delta of each score and the participants', grouped by location, average score, i.e.,\ SVI Q score - average of scores of both images by participants from the same country.
    For example, participants from the US gave an average score of 0.27 to the images from Abuja and Amsterdam, which is 0.06 higher than the score they gave to Abuja and 0.06 lower than the score they gave to Amsterdam.
    The more similar the scores participants give to different scenes, the clearer the color gradient.
    }
    \label{fig:demo-diff}
\end{figure}

\autoref{fig:demo-diff} illustrates differences in perceptions among participants from the same country when evaluating imagery from two different cities.
Although both urban scenes contained very little greenery (GVI $\approx 0.04$), participants disagreed on which was perceived as greener. 
Participants from the USA perceived the image from Abuja to be more \textit{green}, while those from the Netherlands and Chile perceived the opposite, favoring the image from Amsterdam.
In contrast, participants from Singapore and Nigeria considered both images comparable in perceived greenery.
This finding suggests that, while subjective evaluation offers benefits such as providing a more holistic assessment through the lens of lived experiences~\citep{Chaney.2024ujk}, perceived greenery should be considered with caution, as location-specific norms and culture contexts strongly influence such evaluations.

\subsection{Recommendations}
Based on our findings, we have the following recommendations when assessing street-level greenery:

\textbf{GVI alone is not sufficient for evaluating greenery in human-centric applications}. 
Our findings obtained from multiple, diverse cities affirm existing work highlighting that GVI values are not strongly correlated with perceived greenery.
The former consistently underestimates the latter.
In addition to the recommendations on GVI usage highlighted in~\citep{Torkko.2023}, researchers should be aware that relying solely on these measurements shifts the assessment away from a human-centric perspective that accounts for cultural, environmental, and experiential factors.

\textbf{Subjective measurement of greenery is valuable but global models might miss location-specific responses}.
Our findings show that perceived greenery varies the most depending on the participants' location, which remains the only demographic factor that shapes perception of greenery. 
To ensure trustworthy results that reflect the perception of the local population, subjective assessment of greenery should be conducted by residents instead of global model predictions, i.e.,\ prediction models trained on aggregated responses that are usually based on a geographically diverse pool of participants.

\textbf{Arrangement over proximity in perceived greenery}.
Differences in perceived greenery were mainly found when imagery was grouped based on its greenery spatial arrangements rather than when imagery was grouped based on the vegetation distance from the viewer.
Although this pattern may indicate that spatial arrangement influences subjective evaluations, our findings do not reflect participants' explicit preferences for particular configurations. 
Rather, they highlight a direction in which perceived greenery may be especially sensitive. 
Further work that directly examines preference for specific vegetation arrangements, as well as the type of greenery~\citep{torkko2025, korpilo2025}, would help clarify which aspects of greenery most strongly shape subjective assessments.

\subsection{Limitations and prospects for future work}
Subjective perception Q scores and objective GVI measurements each have distinct data requirements that may limit their applicability across different urban contexts. 
Moreover, despite their widespread adoption, SVI is not a panacea and faces inherent limitations in data coverage and representativeness that could affect street-level greenery measurements~\citep{Fan.2025}. 
We inherit some of these limitations from our dataset~\citep{Quintana.2025}. 
Since the dataset was initially designed to evaluate ten perceptual indicators rather than focusing solely on greenery, the surveys were not optimized for demographic comparisons specific to greenery perception. 
Moreover, relative scores are commonly used to assess urban visual perception, making them susceptible to distribution shifts when calculated based on a different pool of participants.
This means that while it is possible to find images with GVI values of 0 or 1, it is almost impossible to find images with a 0 or 1 \textit{greenery} perceived Q score or Trueskill score.
Additionally, while the image and city selection were stratified and conducted at a global scale, it did not explicitly focus on cities with different vegetation types.
Although most participants (78\%) had lived in their respective city for more than five years -- suggesting familiarity with the city's broader environmental and urban-form context -- intra-city variation in greenery can be substantial.
Participants may therefore reside in neighbourhoods that are significantly greener or less green than the citywide average, which could influence their perceptual baselines.
To address this limitation, future work should examine greenery differences within countries and cities, particularly those that are geographically large or environmentally diverse
countries.
The imagery was limited to specific weather conditions and times of day, which standardized the analyses but precluded any temporal or seasonal aspects of greenery evaluation.
Because participants were not provided with a definition of the perceptual\textit{green} indicator, their assessments likely reflected a holistic appraisal that integrated vegetation coverage alongside contextual cues such as colour tone, spatial arrangement, and visible sky. 
Future research should test the robustness of perceptual–objective relationships by analysing images at different seasonal conditions and climate contexts~\cite{torkko2025, zhao2025}.

Regarding the SVI selection within each city, our stratified sampling supports demographic representativeness within cities, but does not capture the geographical diversity needed to compare suburban, urban, or distinct land-use categories. 
Analyses on subjective and objective discrepancies across different urban contexts (e.g.,\ urban vs. suburban areas or by land-use type) is a valuable direction for future research.

Within the images themselves, the vegetation color and saturation could impact GVI values.
For example, trees without leaves in winter may still be classified as green after segmentation, but can differ significantly from a perceived score (e.g.,\ Figure 1 in~\citep{zhao2025}).
Moreover, our GVI measurements were further constrained by the semantic segmentation model's performance and capabilities. 
Using models trained on different datasets could potentially produce varying segmentation results and thus different GVI values~\citep{torkko2025}. 
However, previous research across multiple datasets often reports a similar trend of GVI underestimating perceived greenery~\citep{Torkko.2023, leslie2010, falfan2018}, suggesting this limitation may be systematic rather than dataset-specific.

As highlighted by~\citet{torkko2025}, future work should consider vegetation type diversity to analyze the differential impact of various forms of vegetation on perception.
This vegetation type and spatial arrangements could be part of a modified GVI that could bridge the gap between objective and subjective greenery.
Promising research directions include longitudinal analysis through time-series street view imagery to examine greenery changes in urban development~\citep{liu2025d}, seasonal variation studies to capture temporal dynamics of vegetation perception~\citep{zhao2025}, or expand indices to for easy comparison between regions~\citep{Mahajan.2024}.

\section{Conclusion}\label{sec:conclusion}
Our work advances research on human perception and urban greenery by examining the relationships and differences between perceived and measured greenery across multiple cities and explaining them through human-centric factors such as cultural, environmental, and experiential. 
Conducting a multi-city, demographically diverse survey, we identified a predominantly linear but moderate relationship between perceived and measured street-level greenery, confirming that there is a systematic discrepancy between how green a street feels to people and how green it is actually.

A key result of our study is that most demographic factors make almost no difference in how people perceive greenery, suggesting that survey results may be broadly generalisable and do not necessarily require a demographically balanced sample.
However, we also show that place of residence -- the only demographic factor that showed a measurable influence -- plays a significant role in explaining perceptual variations.
This influence suggests that cultural, environmental, and experiential contexts shape how individuals evaluate urban greenery, underscoring the importance of capturing residents' perceptions when designing such studies.
Our findings also corroborate previous research showing that people's perception of greenery often overestimate objective measurements of greenery.
Image analysis further revealed that residents notice the placement of greenery over mere proximity, suggesting that perception extends beyond simple vegetation presence or coverage. 
This confirms that greenery perception encompasses more than visible vegetation elements and involves nuanced qualitative assessments, not necessarily related to the aesthetics of the urban scene. 

Further research and greenery-focused urban interventions should incorporate both objective and subjective measures, acknowledging their respective strengths and limitations.  
The finding that location trumps proximity suggests that investment in fewer, higher-quality green spaces may be more effective than simply maximizing green coverage, but further work is needed in understanding the type of greenery that influences people's perception the most.

\section*{CRediT authorship contribution statement}
\textbf{M.Q.}:
Conceptualization,
Methodology,
Software,
Validation,
Formal analysis,
Investigation,
Data Curation,
Writing - Original Draft,
Writing - Review \& Editing,
Visualization,
Project administration,
\textbf{F.L.}:
Methodology,
Software,
Data Curation,
Writing - Review \& Editing,
\textbf{J.T.}:
Methodology,
Writing - Review \& Editing,
\textbf{Y.G.}
Methodology,
Writing - Review \& Editing,
Visualization,
\textbf{X.L.}:
Writing - Review \& Editing,
Visualization,
\textbf{Y.H. }:
Data Curation,
Writing - Review \& Editing,
\textbf{K.I.}:
Methodology,
Writing - Review \& Editing,
\textbf{Y.Z.}:
Writing - Review \& Editing,
\textbf{M.A.}:
Writing - Review \& Editing,
\textbf{T.T.}:
Writing - Review \& Editing,
\textbf{Y.L.}:
Methodology,
Writing - Review \& Editing,
\textbf{F.B.}:
Conceptualization,
Methodology,
Resources,
Writing - Review \& Editing,
Visualization,
Supervision,
Funding acquisition.

\section*{Declaration of Competing Interesting}
The authors declare that they have no known competing financial interests or personal relationships that could have appeared to influence the work reported in this article.

\section*{Data availability}
Survey responses and participants' demographic data (SPECS dataset) are released openly (\url{https://github.com/matqr/specs}) and the step-by-step process and code for all analyses are available in the public repository: \url{https://github.com/matqr/greenery-perception}.

\section*{Acknowledgements}
We are grateful for the constructive and encouraging feedback from the editors and reviewers, which has significantly improved our paper.
We thank the participants of the study.
This research was conducted at the Future Cities Lab Global at Singapore-ETH Centre. 
Future Cities Lab Global is supported and funded by the National Research Foundation, Prime Minister's Office, Singapore under its Campus for Research Excellence and Technological Enterprise (CREATE) programme and ETH Z\"urich (ETHZ), with additional contributions from the National University of Singapore (NUS), Nanyang Technological University (NTU), Singapore and the Singapore University of Technology and Design (SUTD).
This research is part of the project Large-scale 3D Geospatial Data for Urban Analytics, which is supported by the National University of Singapore under the Start Up Grant R-295-000-171-133.
This research was supported by the Singapore International Graduate Award (SINGA) scholarship provided by the Agency for Science, Technology, and Research (A*STAR), and the NUS Graduate Research Scholarship (X.L).
This work was supported by the Finnish Ministrys of Education and Culture's Pilot for Doctoral Programmes (Pilot project Mathematics of Sensing, Imaging, and Modelling).
This work was supported by the Research Council of Finland (Flagship of Advanced Mathematics for Sensing, Imaging, and Modelling, FAME, grant number 359182). 
This work was funded by the GREENTRAVEL project by the European Union (ERC, project 101044906). 
Views and opinions expressed are however those of the authors only and do not necessarily reflect those of the European Union or the European Research Council Executive Agency. 
Neither the European Union nor the granting authority can be held responsible for them.

\section*{Declaration of generative AI and AI-assisted technologies in the writing process}
During the preparation of this work, the authors used Claude for proofreading, assistance for code development, and the generation of urban scenes sketches for Figure 1. 
After using this tool/service, the authors reviewed and edited the content as needed and take full responsibility for the content of the published article.

\appendix
 
\clearpage
\section{Methods}\label{app:methods}
\begin{figure}[!ht]
    \centering
    \includegraphics[clip, trim=0cm 3cm 9cm 0cm, width=1\linewidth]{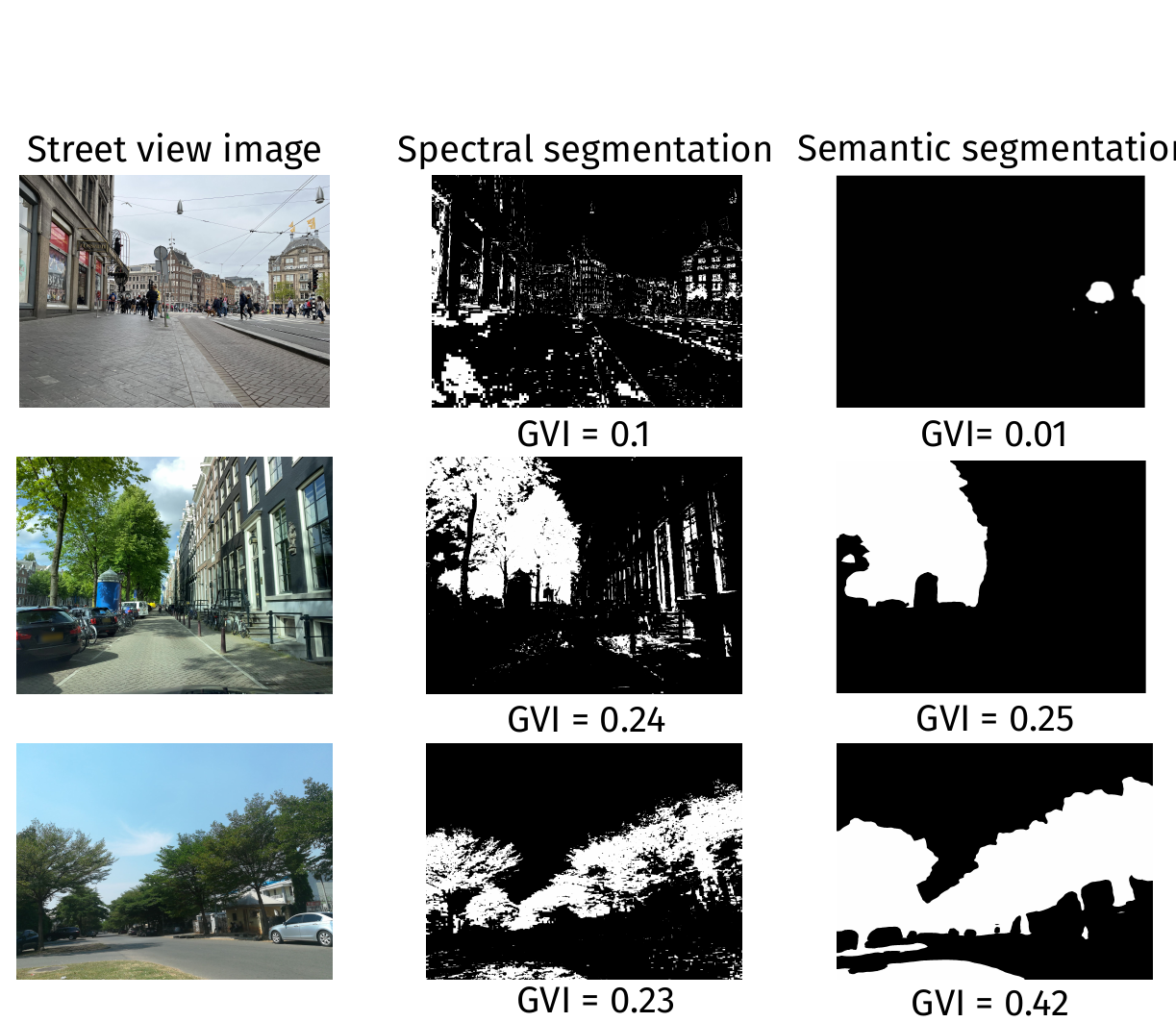}
    \caption{
    Comparison of using spectral and semantic segmentation to mask out pixels in the street view image that represent vegetation.
    The Green View Index (GVI) is calculated following \autoref{eq:gvi} on each segmented image.
    Spectral segmentation is more susceptible to noise (middle column, top and bottom row) due to its dependency on pixel color-values and can lead to a 10$\times$ difference in GVI values compared to semantic segmentation (top row).
    }
    \label{app:fig:spectral-vs-semantic}
\end{figure}

\begin{figure}[!ht]
    \centering
    \includegraphics[width=1\linewidth]{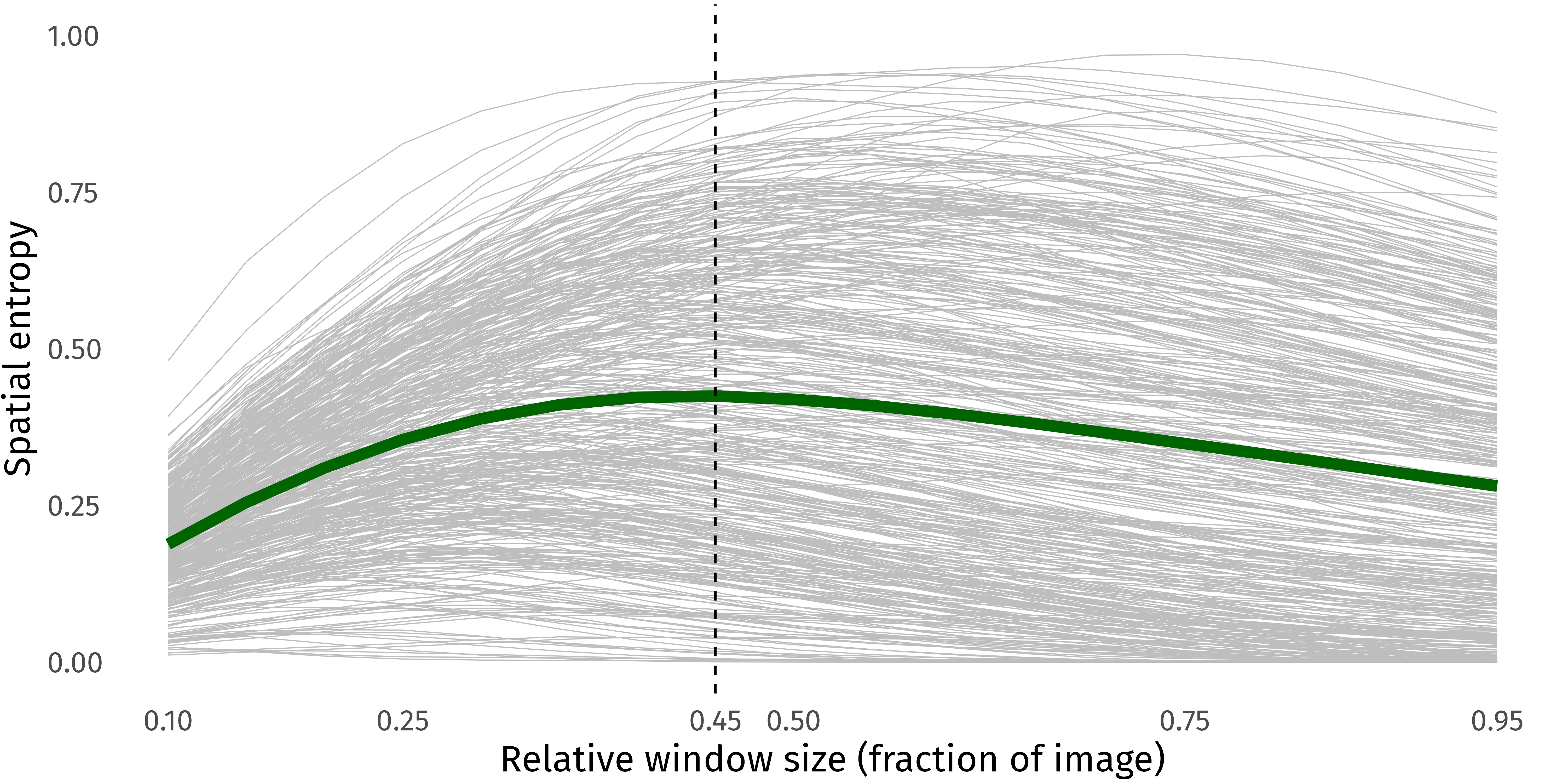}
    \caption{
    Spatial entropy sensitivity analysis with different relative window sizes ($[0.1, 1.0]$ in 0.05 increments) for all 400 images.
    The average line is highlighted in dark green and its maximum entropy is achieved at a relative window size of 0.45.
    }
    \label{app:fig:entropy-sensitivity}
\end{figure}

\begin{figure}[!ht]
    \centering
    \includegraphics[width=1\linewidth]{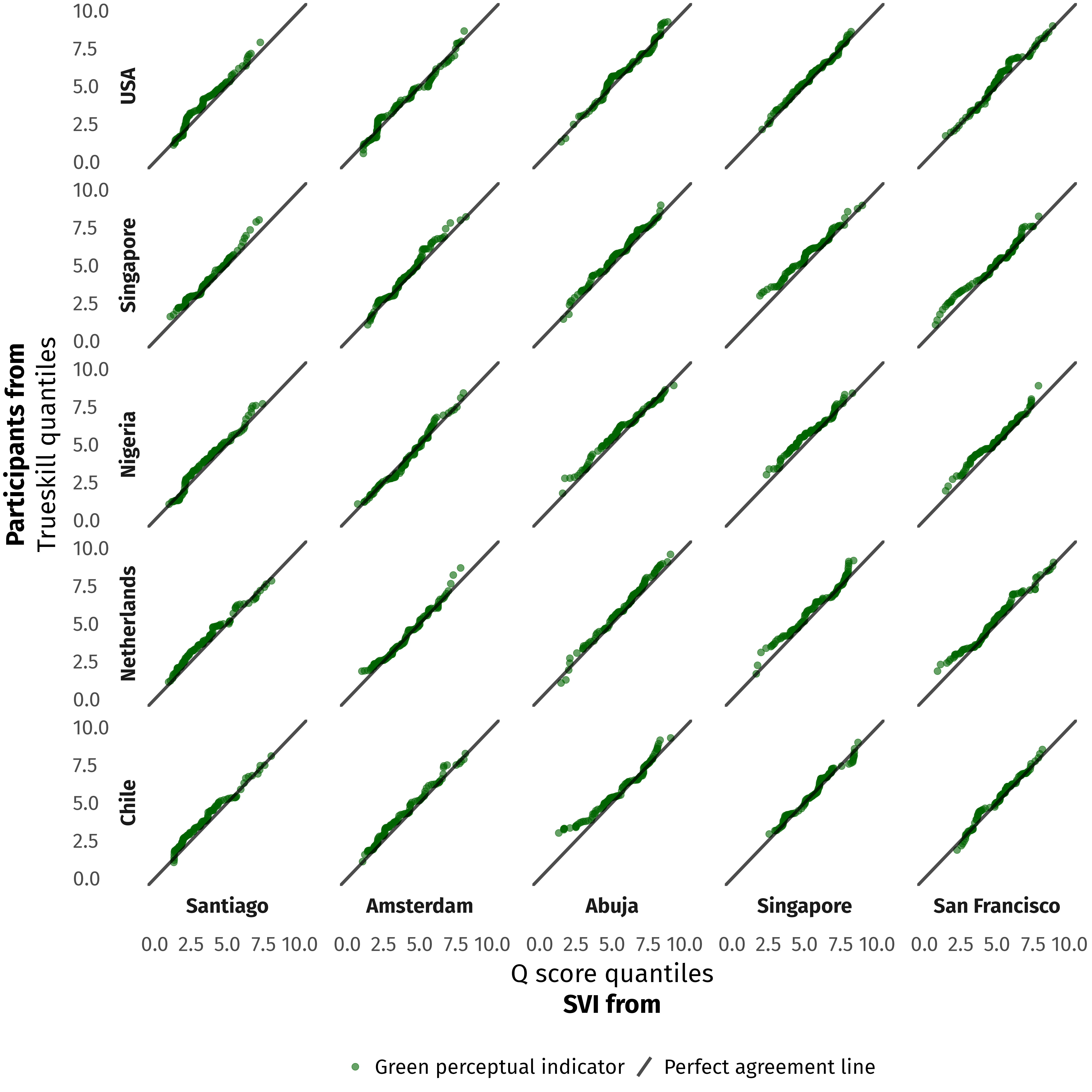}
    \caption{
    Quantile-Quantile comparison between Q scores and Trueskill scores for the \textit{green} perceptual indicators grouped by images' and participants' location pairs (country-city pairs).
    Each location pair has at least 54 images with more than four pairwise comparisons ($n\geq54$).
    }
    \label{app:fig:qscore-trueskill-qq}
\end{figure}

\clearpage

\begin{table}[!ht]
    \centering
    \begin{tabular}{cc}
        \hline
        \multicolumn{2}{c}{\textbf{(a) Hyperparameters}} \\
        \hline
        \textit{n\_estimators} & 100, 150, 200, \dots, 500 \\
        \textit{max\_depth} & 10, 20, 30, None \\
        \textit{min\_samples\_split} & 2, 5, 10 \\
        \textit{min\_samples\_leaf} & 1, 2, 4 \\
        \textit{max\_features} & \textit{sqrt}, \textit{log2}, None\\
        \\
        
        \hline
        \multicolumn{2}{c}{\textbf{(b) Best performing model}} \\
        \hline
        \textit{n\_estimators} & 200 \\
        \textit{max\_depth} & None \\
        \textit{min\_samples\_split} & 2 \\
        \textit{min\_samples\_leaf} & 2 \\
        \textit{max\_features} & None \\
        \textit{MSE$_{cross-validation}$} & 0.0187 \\
        \textit{MSE$_{\text{test set}}$} & 0.0191 \\
        \textit{R$^2$} & 0.49\\
        \hline
    \end{tabular}
    \caption{
    Overview of the Random Forest model training process and results.
    We used a 5-fold cross-validation for 50 candidates, based on a randomized combination of the available \textbf{(a)} hyperparameters, and reported the \textbf{(b)} best performing model based on Mean Squared Error (MSE).
    }
    \label{app:tab:model-training}
\end{table}
\clearpage

\section{Results}\label{app:results}
\begin{figure}[!ht]
    \centering
    \begin{subfigure}{\linewidth}
        \centering
        \includegraphics[width=\linewidth]{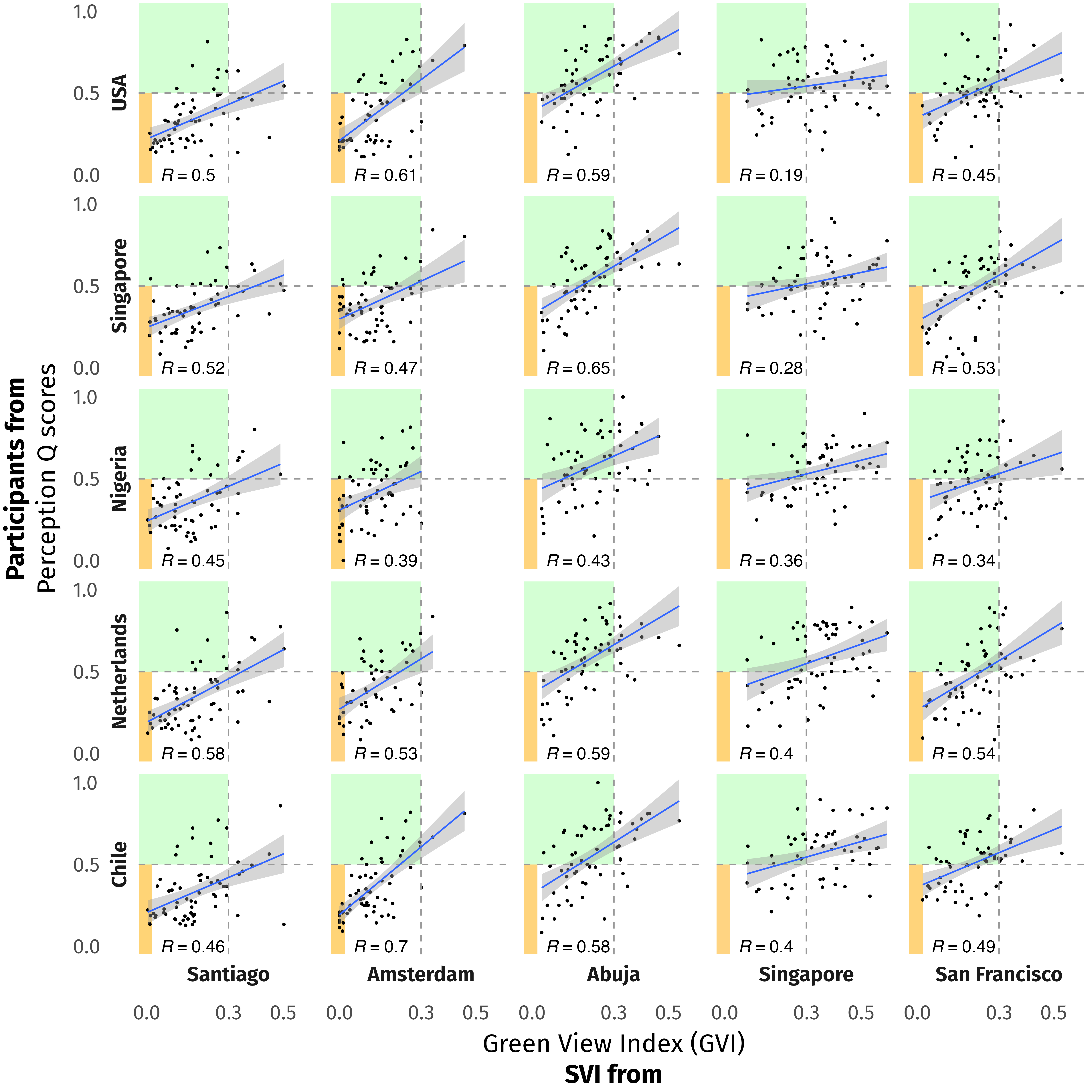}
        \caption{
        \textit{Green} perception Q scores are calculated based on the images' and participants' location pair.
        Each location pair has at least 54 images with more than four pairwise comparisons ($n\geq54$).
        }
        \label{fig:app:green-gvi-corr-scatter_location}    
    \end{subfigure}
\end{figure}
\begin{figure}
    \centering
    \ContinuedFloat
    \begin{subfigure}{\linewidth}
        \centering
        \includegraphics[width=\linewidth]{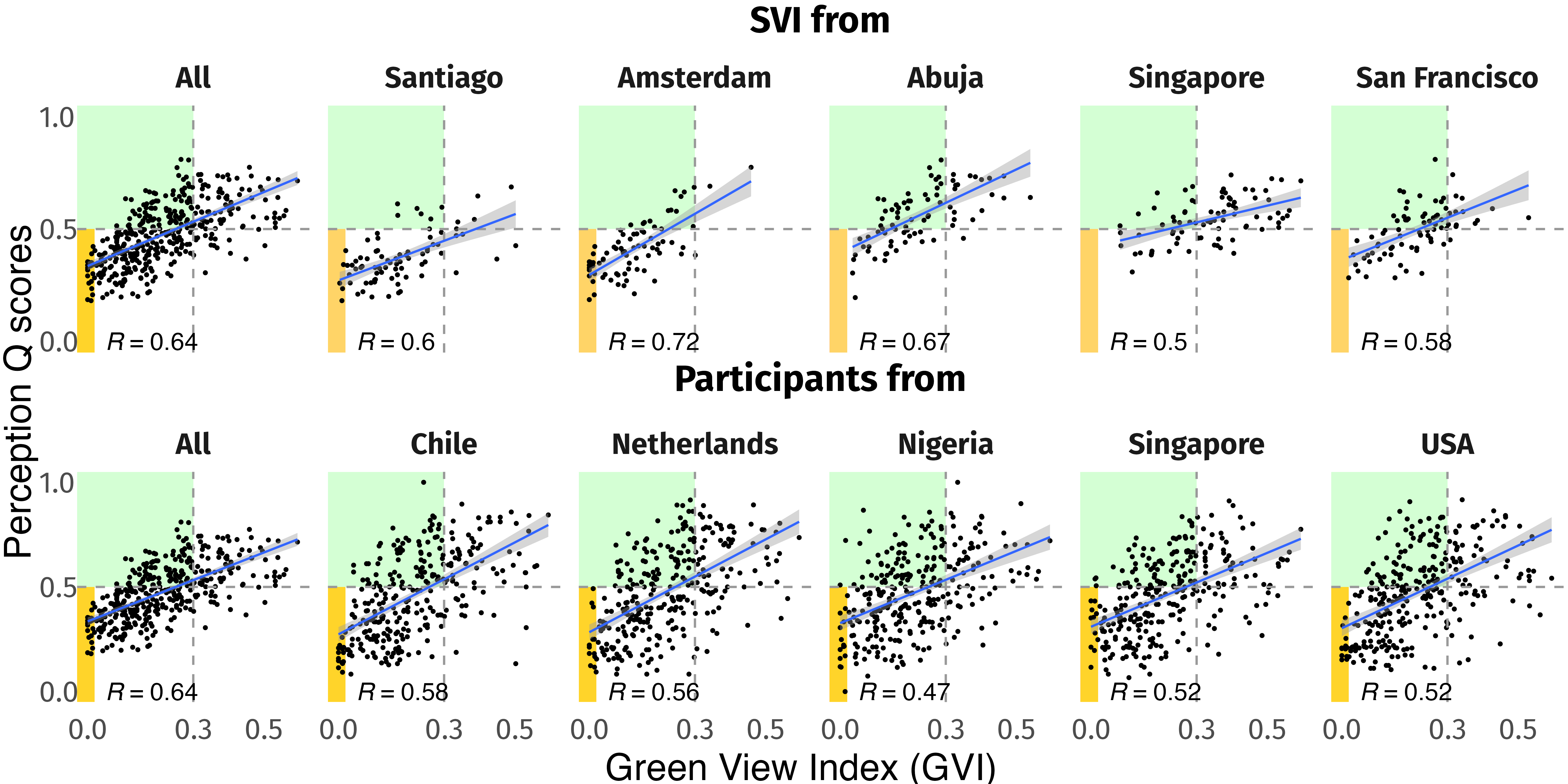}
        \caption{
        \textit{Green} perception Q scores are calculated based on the images' or participants' individual location.
        Each grouping has 80 images with more than four pairwise comparisons ($n = 80$) (top row, `All' group $n = 400$) and at least 297 images with more than four pairwise comparisons ($n \geq 297$) (bottom row).
        }
        \label{fig:app:green-gvi-corr-scatter}
    \end{subfigure}
    \caption{
    Correlation between green perception Q scores and GVI values within images' and participants' location pairs (country-city pairs) (a) and images' or participants' individual location (b).
    Region in green captures the imagery with some vegetation (GVI$<0.3$) perceived moderately green (\textit{green} perception Q score $>0.5$) and region in yellow captures the imagery with very little vegetation (GVI$<0.02$) perceived as not so green (\textit{green} perception Q score $<0.5$).
    }
    \label{fig:app:green-gvi-scatter}
\end{figure}

\begin{figure}[!ht]
    \centering
    \includegraphics[width=\linewidth]{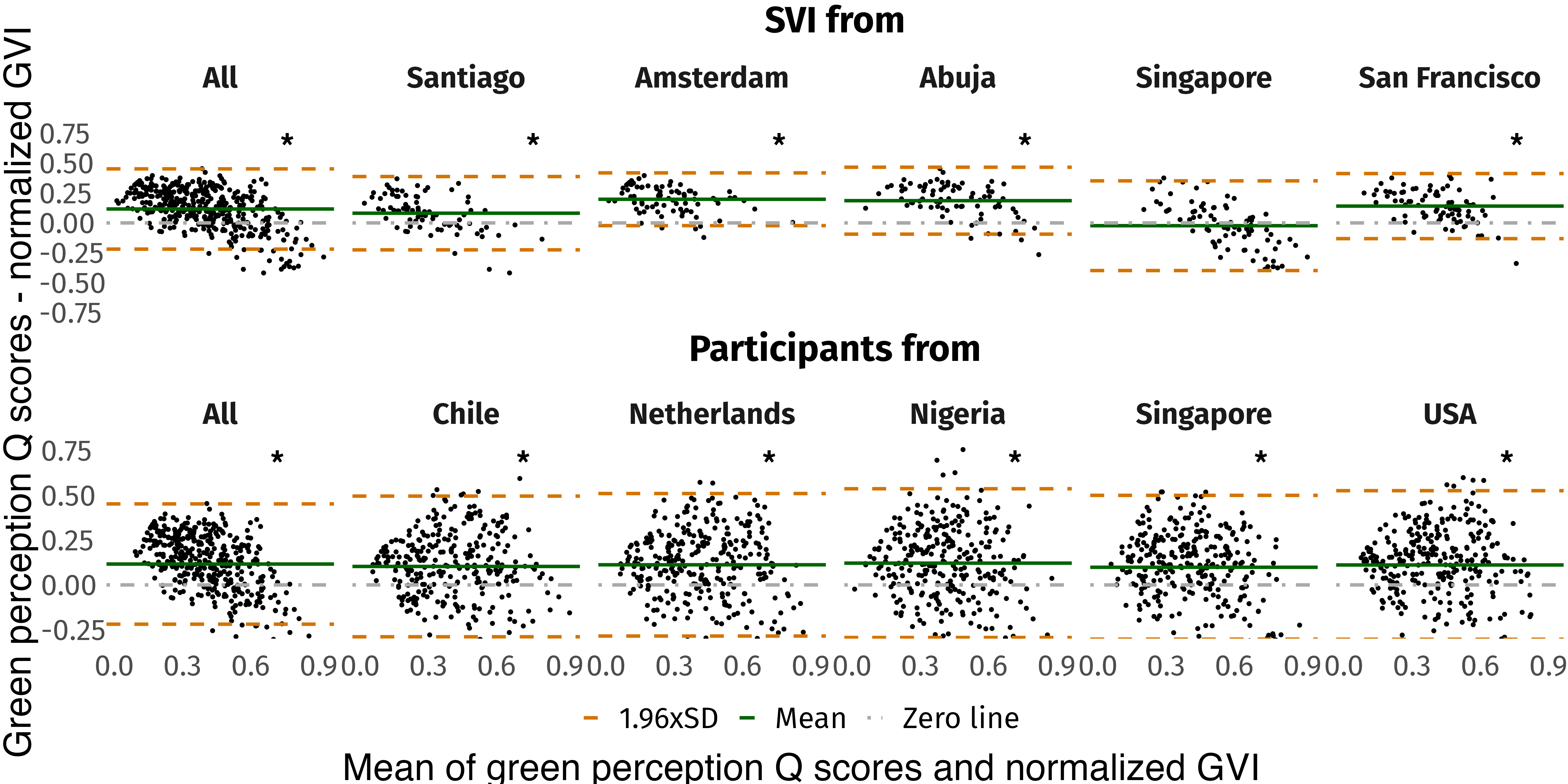}
    \caption{
    Bland-Altman plot comparing differences between \textit{green} perception Q Scores and normalized Green View Index (GVI) values across different location pairs (country-city pairs). 
    The plot shows the difference between measurements (y-axis) against their mean (x-axis) grouped by images' and participants' location, with green horizontal lines representing the mean difference, dashed orange lines indicating $\pm$1.96 standard deviations, and a dotted gray line at zero.
    Each grouping has 80 images with more than four pairwise comparisons ($n = 80$) (top row, `All' group $n = 400$) and at least 297 images with more than four pairwise comparisons ($n \geq 297$) (bottom row).
    Plots with significant differences between green perception Q score and normalized GVI are shown (*$p<0.05$).
    }
    \label{fig:app:bland-altman}
\end{figure}

\begin{figure}[!ht]
    \centering
    \includegraphics[width=\linewidth]{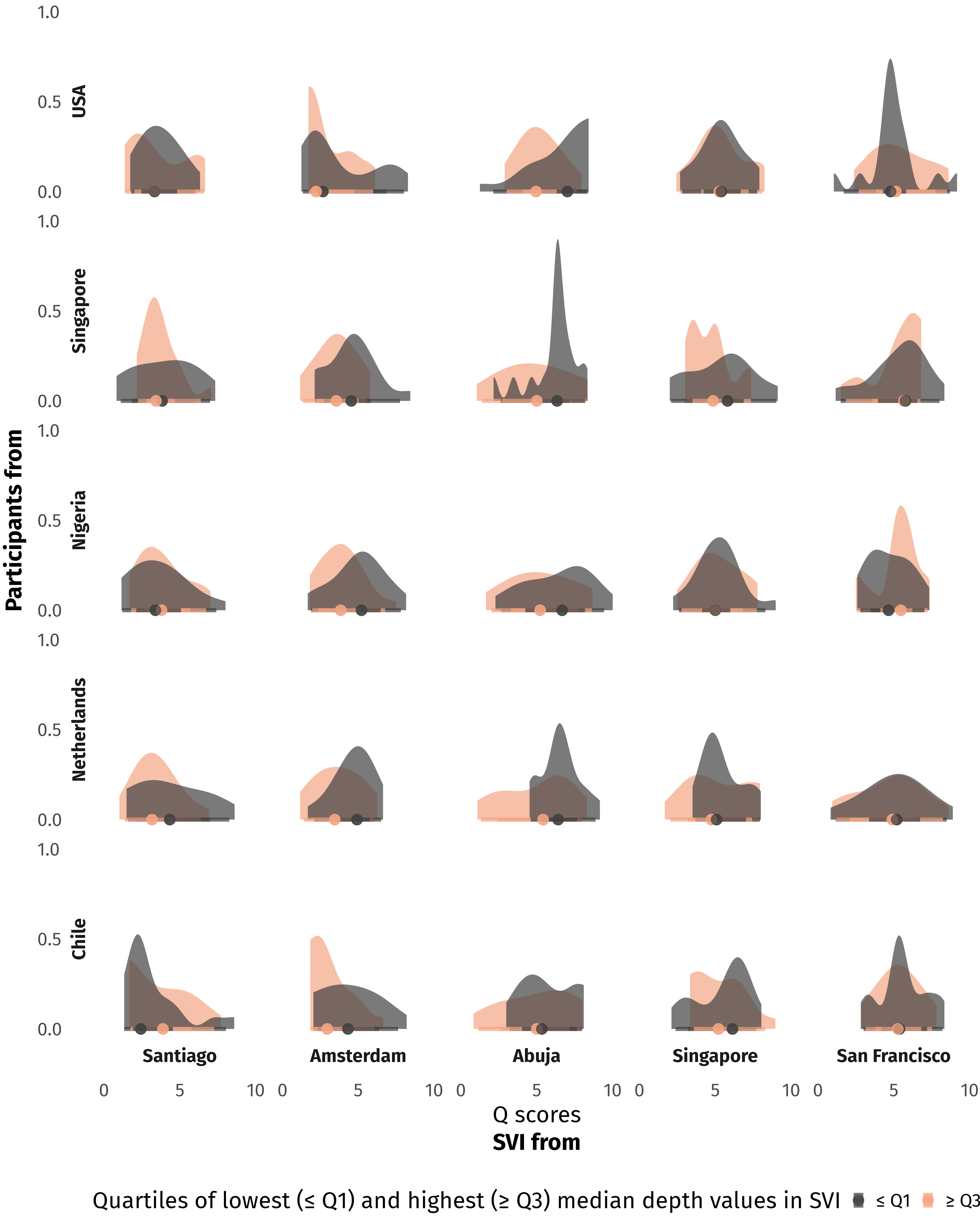}
    \caption{
    \textit{Green} perception Q score distributions of street view imagery (SVI) within the 25$^{th}$ ($\leq$Q1 in dark grey) or within the 75$^{th}$ ($\geq$Q3 in pink) percentile of their vegetation proximity in meters (median of depth values).
    Subjective Q scores are calculated based on the images' and participants' location pairs (country-city pairs), median  represented by a filled circle.
    No significant differences between quantiles were found, and each location pair has more than 14 images with more than four pairwise comparisons ($n\geq14$).
    }
    \label{fig:app:qscore_by_depth_location}
\end{figure}

\clearpage
\bibliographystyle{elsarticle-harv} 
\bibliography{1_green}

\end{document}